\documentclass[journal]{IEEEtran}
\usepackage[font=footnotesize,labelfont=bf]{caption}
\usepackage{mathrsfs}
\usepackage{amsmath,amssymb,amsfonts,bm}
\usepackage{algorithmic}
\usepackage{algorithm}
\usepackage{array}
\usepackage{textcomp}
\usepackage{stfloats}
\usepackage{url}
\usepackage{verbatim}
\usepackage{epstopdf} 
\usepackage{longtable}
\usepackage{float}
\usepackage{graphics}
\usepackage{graphicx}
\usepackage{subcaption}
\usepackage{tabularx} 
\usepackage{booktabs} 
\usepackage{hhline} 
\usepackage[doipre={doi:~}]{uri}
\usepackage{color,xcolor}
\usepackage[numbers]{natbib}
\usepackage{makecell}

\makeatletter
\renewcommand\fs@ruled{%
  \def\@fs@cfont{\bfseries}%
  \let\@fs@capt\floatc@ruled
  \def\@fs@pre{\hrule height\heavyrulewidth\kern2pt}%
  \def\@fs@mid{\kern2pt\hrule height\lightrulewidth\kern2pt}%
  \def\@fs@post{\kern2pt\hrule height\heavyrulewidth}%
  \let\@fs@iftopcapt\iftrue
}
\floatstyle{ruled}
\restylefloat{algorithm}
\makeatother

\usepackage{multicol}

\begin{document}

\title {Cloud Workflow Scheduling Based on Graph Attention-Driven Hierarchical Reinforcement Learning}

\author{Zongjin Li, Shaohan Feng~\IEEEmembership{Member,~IEEE}, Chunxi Yang~\IEEEmembership{Member,~IEEE}  and Wenbo Wang {\emph{Senior Member, IEEE}}, 
\thanks{Zongjin Li, Chunxi Yang and Wenbo Wang are with the Faculty of Mechanical and Electrical Engineering, Kunming University of Science and Technology, Kunming 650500, China (e-mails: zongjin\_li@stu.kust.edu.cn;  ycx@kust.edu.cn; wenbo\_wang@kust.edu.cn). 

Shaohan Feng is with the School of Information and Electronic Engineering
(Sussex Artificial Intelligence Institute), Zhejiang Gongshang University,
Hangzhou 310018, China (e-mail: feng\_shaohan@mail.zjgsu.edu.cn).
}
}

\maketitle

\begin{abstract}
Dynamic cloud workflow scheduling must balance deadline satisfaction, container utilization, and energy consumption while dealing with stochastic task-execution speeds, placement-dependent communication, and coupled task and container decisions. Workflows are naturally modeled as directed acyclic graphs (DAGs), but conventional vector- or matrix-based states do not fully capture their dependency topology. To better represent task urgency and structural relationships, we assign predicted sub-deadlines to tasks and use a multi-head graph attention network (GAT) to extract dependency information from the evolving DAGs. Based on these representations, we develop a Graph Attention-Driven Hierarchical Reinforcement Learning (GA-HRL) framework and model the scheduling process as an event-driven hierarchical semi-Markov decision process (SMDP). Workflow arrivals and task completions trigger scheduling events. At each scheduling event, the Task Scheduling (TS) agent first processes the currently ready tasks by assigning them to admissible existing containers or requesting new ones. The requested containers are then processed by the Container Scheduling (CS) agent for host placement before the environment advances. The two agents are trained alternately using separate Proximal Policy Optimization (PPO). Experiments on the 2018 Alibaba cluster trace show that GA-HRL maintains competitive workflow success rate and, in settings where success is comparable, generally achieves higher container utilization and lower energy consumption. Under the largest speed variation, it trades a small success-rate margin for substantially lower energy. Simulation code is available at: \url{https://github.com/zongjin130/GA-HRL}.
\end{abstract}


\section{Introduction}
\IEEEPARstart{I}{n} cloud computing environments, users submit their computational demands as workflows to the cloud platform, and the provider must allocate computing resources according
to each workflow's task structure and dependencies~\cite{9491116}. The scheduler therefore has to balance multiple objectives simultaneously~\cite{8443134}. From the user perspective, workflows should finish before their deadlines to maintain quality of service~\cite{cheng2015energy}. From the provider perspective, allocated resources should be utilized effectively, because unnecessary provisioning and poor utilization keep additional hosts and containers active and increase energy consumption. Consequently, dynamic workflow scheduling must jointly consider task timeliness, resource utilization, and energy efficiency rather than optimizing them separately.

Cloud workflows are becoming larger and more complex, and cloud
platforms usually operate online, so multiple workflows with different structures, arrival times, and deadlines may coexist~\cite{8443134}. Scheduling such workflows is difficult for two related reasons. First, a workflow is successful only after all its constituent tasks are coordinated, so task decisions are coupled. Second, precedence constraints restrict
the order of execution, and treating tasks independently discards
useful dependency information. Workflows are therefore commonly
represented as Directed Acyclic Graphs (DAGs), where nodes denote
tasks and directed edges represent both precedence constraints and
data transfers~\cite{10497174}.

From the resource-management perspective, containers are widely used because they are lightweight, start quickly, and support high resource density~\cite{10499978}. The platform may also need to create additional containers on suitable hosts when current resources are insufficient to meet workflow deadlines~\cite{9991105}. In practice, however, container performance is not fully predictable: the execution speed of a task can vary with host load and placement-dependent interference~\cite{8443134}. Communication is likewise placement-dependent, because intra-host and inter-host transfers use different bandwidths and transfer time cannot be known until containers are placed~\cite{cheng2015energy}. These uncertainties interact with workflow dependencies and can cause waiting time, reduce container reuse, and increase energy consumption. Dynamic workflow scheduling therefore requires a model that captures both runtime uncertainty and placement-dependent communication.

Cloud workflow scheduling requires trade-offs both among tasks of the same workflow and among workflows competing for shared resources, and the problem is NP-hard~\cite{topcuoglu2002performance,10061217}. Traditional heuristics depend on problem-specific rules, whereas optimization-based methods often rely on repeated search or expensive solving procedures. As the numbers of tasks and candidate resources grow, such methods become difficult to deploy in an online scheduler. Reinforcement Learning (RL) has therefore attracted attention as a way to learn scheduling policies from interaction with the environment~\cite{mnih2015human}. However, many RL-based schedulers still represent tasks by ordinary vectors or matrices, which retain local attributes but do not explicitly capture the non-Euclidean dependency structure of a DAG~\cite{10463608}. How to preserve and exploit this structure under changing resource conditions remains an open problem.

This paper studies dynamic cloud workflow scheduling with stochastic task-execution speeds and placement-dependent communication. Each active workflow is kept as a DAG, and a predicted sub-deadline is assigned to every task to represent its temporal urgency. A multi-head GAT then aggregates dependency information so that tasks with similar local attributes can still be distinguished by their structural context. The resulting representation is used in an event-driven hierarchical semi-Markov Decision Process (SMDP), where a Task Scheduling (TS) agent first assigns ready tasks to admissible existing containers or requests new ones. All new-container requests generated during the TS phase are then processed by a Container Scheduling (CS) agent for host placement before physical time advances. The two policies are trained alternately using separate PPO actor-critic networks while sharing the same scheduling environment. GA-HRL is designed to meet workflow deadlines while also improving container utilization and reducing total energy consumption. The main contributions are as follows:

\begin{itemize}
\item [1)] We develop a dependency-aware task representation for dynamically arriving DAG workflows. Predicted sub-deadlines express task urgency, while a multi-head GAT aggregates topological information that conventional vector or matrix states do not retain explicitly.

\item [2)] We formulate the coupled task-to-container and container-to-host decisions as an event-driven hierarchical SMDP. At each scheduling event, the TS agent first processes the ready-task assignments, after which the CS agent processes the host-placement decisions generated by new-container requests before physical time advances.

\item [3)] We train the two agent-specific policies with PPO over
their event-driven decision sequences. Trace-driven experiments show that GA-HRL maintains competitive workflow success and, in settings where success is comparable, achieves higher container utilization and lower energy consumption. Under the largest speed variation, it trades a small success-rate margin for substantially lower energy.
\end{itemize}

The remainder of this paper is organized as follows. Section~\ref{lab-related-work} reviews related work. Section~\ref{lab-sec-model} introduces the system model and problem statement. Section~\ref{lab-sec-alg} presents the proposed GA-HRL scheduling algorithm. Section~\ref{lab-sec-experiment} reports the experimental results, and Section~\ref{lab-sec-conclusion} concludes the paper.

\section{Related Works}
\label{lab-related-work}
This section reviews representative studies on cloud workflow scheduling from four perspectives: containerized workflow scheduling, mathematical optimization methods, heuristic optimization methods, and machine-learning-based methods. The discussion focuses on how these approaches represent workflow dependencies, handle resource allocation, and respond to dynamic or uncertain execution conditions.

\subsection{Containerized Workflow Scheduling}

Workflows are commonly represented by Petri nets~\cite{kheldoun2017formal}, UML diagrams, or
DAGs~\cite{10497174}. DAGs are widely used because they naturally express task parallelism, precedence constraints, and data dependencies. Each workflow DAG has its own arrival time, structure, and deadline~\cite{8443134}.

In containerized cloud environments, the scheduler must make two
coupled decisions: task scheduling, which determines the execution
order and start time of tasks, and container placement, which
determines the host on which each container is deployed~\cite{9991105}. These two decisions jointly affect execution efficiency and resource consumption. In practice, container execution speed fluctuates with host load, and communication delay depends on whether containers are co-located. These uncertainties interact with workflow dependencies, which can waste resources and increase energy consumption.

\subsection{Mathematical Optimization based Methods}

Optimization-based methods formulate cloud resource allocation or
workflow scheduling as mathematical programs. Chen \emph{et al.}~\cite{cheng2015energy} reconstructed the request sequence by priority and minimized matching distance and the number of active physical machines. Jiao \emph{et al.}~\cite{jiao2017joint} formulated joint resource placement and allocation as an integer linear program and designed a dynamic-programming-based algorithm. Hahnel \emph{et al.}~\cite{hahnel2018extending} extended the cutting-stock model to consolidate heterogeneous service requests,
reducing overload probability and energy consumption. Liu \emph{et al.}~\cite{liu2019dependent} proposed an approximate function placement algorithm for edge-cloud job completion time minimization, and Das \emph{et al.}~\cite{das2020performance} studied dynamic function placement under cost and deadline constraints. Deng \emph{et al.}~\cite{deng2021dependent} later
embedded dependent functions into distributed serverless edge
computing to obtain the optimal function placement and start time.

These methods provide clear formulations and, in some cases,
approximation guarantees. However, they generally assume deterministic inputs or require repeated optimization as the system state changes. This limits algorithm scalability because the computational burden grows quickly with the decision space. In our setting, workflows arrive online, execution speeds are stochastic, and each task assignment affects subsequent container placement, making direct online application of such methods difficult.

\subsection{Heuristic Optimization based Methods}

Heuristic methods can be broadly divided into rule-based scheduling
and iterative swarm-intelligence methods~\cite{8443134}. Rodriguez \emph{et al.}~\cite{rodriguez2014deadline} used particle swarm optimization to maximize the scheduling success rate and minimize cost in static cloud workflows. Arabnejad \emph{et al.} \cite{arabnejad2017scheduling} proposed deadline-based heuristics for dynamic cloud workflows. Chen \emph{et al.} \cite{8443134} developed an uncertainty-aware online scheduler for real-time workflows with multiple objectives, while Fan \emph{et al.} \cite{9991105} proposed an energy-efficient heuristic for deadline-constrained workflows with container placement.

These heuristics are generally computationally efficient and often work well under fixed assumptions. However, their performance depends heavily on hand-crafted rules or tuned parameters. When the arrival process, execution-speed distribution, or resource configuration changes, the same rules may not adapt as effectively as a learned policy.

\subsection{Machine Learning based Methods}

Machine-learning-based schedulers can be divided into methods that use learning as a predictor/evaluator and methods that use RL to generate policies~\cite{yang2024scheduling, 10634877}. Yang \emph{et al.} \cite{yang2024scheduling} combined machine-learning prediction with relaxed linear programming to schedule tasks with unknown execution times. Yu \emph{et al.}~\cite{10061217} applied RL with custom reward functions to optimize dynamic workflow scheduling. Ding \emph{et al.} \cite{10634877} proposed a Transformer-enhanced Deep Q-Network for large-scale workflow scheduling, but the approach does not fully exploit inter-task dependency structure. Xie \emph{et al.} \cite{9590522} used graph neural networks to extract features of workflows and resources, improving scheduling success and energy efficiency.

Despite these advances, many existing schedulers use dependency
information mainly to determine eligible tasks, rather than embedding the DAG topology directly into the policy state. As a result, the structural information carried by the DAG is only partially exploited when resources are selected. Moreover, several methods assume that workflows are available at the beginning of scheduling or use pre-execution attributes, which does not fully capture the continual changes caused by online arrivals, fluctuating execution speeds, and dynamic container availability.

GA-HRL addresses these gaps by combining DAG-preserving GAT representations, predicted sub-deadlines, and a hierarchical TS/CS policy in an event-driven scheduling framework. It differs from prior cloud workflow schedulers at three connected levels:
\begin{itemize}
  \item it explicitly models random workflow arrivals, task-specific execution-speed realizations, and placement-dependent communication;
  \item at the TS level, the task state combines a predicted urgency indicator with a GAT representation of the DAG topology; and
  \item for policy derivation, the state drives an event-triggered hierarchical policy that coordinates task assignment and container placement at separate decision epochs.
\end{itemize}

\section{Model and Problem Statement}
\label{lab-sec-model}
This section focuses on three key aspects: cloud resource modeling, workflow modeling, and the problem statement. 

\subsection{Cloud Resource Model}

We consider a cloud service center in which physical hosts are activated on demand. Let $\mathcal H=\{H_1,H_2,\cdots,H_N\}$ denote the set of host instances activated during a scheduling episode, where $N$ is the total number of activated host instances. Each host $H_n$ is characterized by a resource tuple $H_n=(C_n,M_n,\bar Q_n,\bar P_n)$, where $C_n$ represents the number of CPU cores, $M_n$ is the memory size, $\bar Q_n$ is the mean computational capacity measured in Million Instructions Per Second (MIPS), and $\bar P_n$ is the mean power under full utilization. Let $\mathcal C^{\mathrm{all}}$ denote the set of all containers created during the scheduling horizon, and let $\mathcal C_n(t)$ denote the set of containers that are currently alive on host $H_n$ at time $t$. We further define $\mathcal C_n^{\mathrm{all}}=\{c_m\in\mathcal C^{\mathrm{all}}:\eta_m=n\}$ as the set of all containers deployed on $H_n$ during the scheduling horizon. Then, at any time, a container $c_m\in\mathcal C_n(t)$ is characterized by the resources allocated to it: $c_m=(C_m,M_m)$, where $C_m$ and $M_m$ are respectively the number of CPU cores and memory required for container $c_m$.

The mean computational capacity and mean power of the container are determined by the number of
CPU cores. The mean computational capacity of container $c_m$ is expressed as
\begin{equation}
\bar Q_m=\frac{\bar Q_{\eta_m}C_m}{C_{\eta_m}},
\end{equation}
where $\eta_m=n$ means that container $c_m$ is deployed on host
$H_n$.
The nominal capacity $\bar Q_m$ is known to the scheduler, whereas the capacity realized during execution is task specific. Let $Q_{k,j}^{(m)}$ denote the computational capacity experienced by task $t_{k,j}$ when it executes on container $c_m$. At the start of every task execution, a new realization is sampled independently as
\begin{equation}
Q_{k,j}^{(m)}\sim\mathcal{N}\!\left(\bar Q_m,\,(v\bar Q_m)^2\right),
\;\; 0<Q_{k,j}^{(m)}<2\bar Q_m,
\label{eq:speed}
\end{equation}
where $v$ is the variance coefficient. 
Thus, two tasks executed successively on the same container may experience different realized capacities, while scheduling decisions made before execution use the nominal value $\bar Q_m$.
We assume that the data transmission speed between containers is different \cite{9991105}: the
communication speed within the same host shares the internal network bandwidth of the host,
usually with lower latency and not limited by physical networks. The communication speed in this
scenario is denoted as $B^{in}$. $B^{in}$ represents the effective intra-host bandwidth under the adopted sharing assumption. When the container is deployed on different hosts, all data
transmission must pass through a shared physical network link, and its bus bandwidth, denoted as $B^{cr}$, is strictly
limited by the physical bandwidth of the cross-host link.

\subsection{Workflow Model}
We consider a set of $\mathcal W=\{W_1,\ldots,W_K\}$ of $K$ workflows.
Each workflow is characterized by the tuple $W_k=(A_k,D_k,G_k)$, where
$A_k$ is its arrival time, $D_k$ is its deadline, and
$G_k=(\mathcal T_k,\mathcal E_k)$ is the DAG representing its task structure. In particular, $\mathcal T_k=\{t_{k,1},\ldots,t_{k,N_k}\}$ is the task set of $W_k$, with $N_k=|\mathcal T_k|$, and
$\mathcal E_k\subseteq\{e_{k,ij}\mid i,j\in\{1,\ldots,N_k\},\ i\ne j\}$
is the edge set representing the data dependency between tasks. An edge $e_{k,ij}$ indicates that $t_{k,i}$ is an immediate predecessor of $t_{k,j}$, equivalently, $t_{k,j}$ is an immediate successor of $t_{k,i}$. We denote the immediate predecessor and successor sets of $t_{k,j}$ by $\operatorname{pred}(t_{k,j})$ and $\operatorname{succ}(t_{k,j})$, respectively. A task with no immediate predecessor is called an \emph{entry} task, and a task with no immediate successor is called an \emph{exit} task. Each task $t_{k,j}$ has a required computation size $p_{k,j}$, measured in Million Instructions, and each edge $e_{k,ij}$ carries
a data volume $d_{k,ij}$ that must be transmitted from $t_{k,i}$ to
$t_{k,j}$. We assume that the workflow structure, task computation
sizes, and inter-task data volumes are known when $W_k$ arrives (see also~\cite{9491116}).

\begin{figure}[!t]
    \centering
    \includegraphics[width=3.4 in]{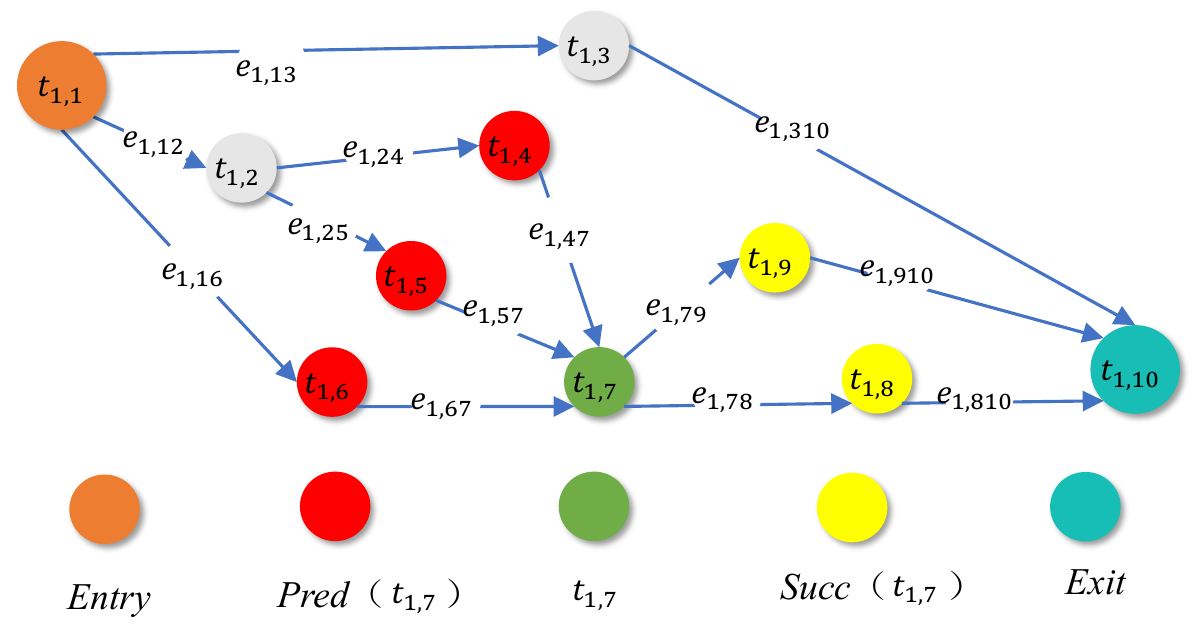}
    \caption{Workflow diagram.}
    \label{Fig_0}
\end{figure}

An illustrative example is shown in Fig \ref{Fig_0}, where tasks $t_{1,1}$ and $t_{1,10}$ are respectively the \emph{entry} task and the \emph{exit} task. Tasks $t_{1,4}$, $t_{1,5}$, and $t_{1,6}$ are the immediate predecessors of task $t_{1,7}$. Tasks $t_{1,8}$ and $t_{1,9}$ are the immediate successors of task $t_{1,7}$.

We introduce two mappings for notational convenience: $\mu_{k,j}=m$ (task $t_{k,j}$ assigned to container $c_m$) and $\eta_m=n$ (container $c_m$ deployed on host $H_n$). These are merely index simplifications and leave the scheduling model unchanged. The main notations are summarized in Table~\ref{tab:main_notation}.

\begin{table}[!t]
\centering
\caption{Main notation used in the scheduling model.}
\label{tab:main_notation}
\scriptsize
\renewcommand{\arraystretch}{1.08}
\begin{tabularx}{\columnwidth}
{@{}>{\raggedright\arraybackslash}p{0.34\columnwidth}X@{}}
\toprule
Symbol & Meaning \\
\midrule

$\mathcal H,H_n$ &
set of host instances and host instance $n$ \\

$C_n,M_n$ &
CPU-core capacity and memory capacity of host $H_n$ \\

$\bar Q_n,\bar P_n$ &
mean computational capacity and mean full-utilization power of host $H_n$ \\

$\mathcal W,W_k$ &
set of workflows and workflow $k$ \\

$A_k,D_k$ &
arrival time and deadline of workflow $W_k$ \\

$G_k=(\mathcal T_k,\mathcal E_k)$ &
DAG of workflow $W_k$, consisting of task and edge sets \\

$N_k$ &
number of tasks in workflow $W_k$ \\

$t_{k,j},e_{k,ij}$ &
task $j$ of workflow $W_k$ and dependency edge from $t_{k,i}$ to $t_{k,j}$ \\

$\operatorname{pred}(t_{k,j}),\operatorname{succ}(t_{k,j})$ &
immediate predecessor and successor sets of task $t_{k,j}$ \\

$p_{k,j},d_{k,ij}$ &
computational workload of task $t_{k,j}$ and data volume transmitted from $t_{k,i}$ to $t_{k,j}$ \\

$\mathcal C^{\mathrm{all}},c_m$ &
set of all containers created during the scheduling horizon and globally indexed container $m$ \\

$\mathcal C_n(t)$ &
set of containers currently alive on host $H_n$ at time $t$ \\

$\mathcal C_n^{\mathrm{all}}$ &
set of all containers deployed on host $H_n$ during the scheduling horizon \\

$C_m,M_m$ &
CPU-core and memory requirements of container $c_m$ \\

$\bar Q_m,Q_{k,j}^{(m)}$ &
nominal capacity of container $c_m$ and capacity realized by task $t_{k,j}$ on $c_m$ \\

$\bar P_m,P_{k,j}^{(m)}$ &
nominal container power and task-specific power during execution \\

$\mu_{k,j}$ &
index of the container assigned to task $t_{k,j}$ \\

$\eta_m$ &
host index of container $c_m$; $\eta_m=n$ iff $c_m\in\mathcal C_n^{\mathrm{all}}$ \\

$B^{in},B^{cr}$ &
intra-host and cross-host data transmission bandwidths \\

$n_{\mathrm{act}}^{n}(t)$ &
number of containers on host $H_n$ simultaneously receiving data at time $t$ \\

$ST_c$ &
startup time required to deploy a new container \\

$\tau^{\mathrm{ex}}_{k,j}$ &
execution time of task $t_{k,j}$ \\

$\tau^{\mathrm{tx}}_{k,ij}$ &
data transmission time from task $t_{k,i}$ to task $t_{k,j}$ \\

$R_m,S_{k,j},F_{k,j},F_k$ &
container-ready, task-start, task-finish, and workflow-finish times \\

$F_m^{\mathrm{cur}}$ &
scheduled finish time of the task currently executing on container $c_m$ \\

$T_n^{\mathrm{on}}$ &
total powered-on duration of host $H_n$ \\

$T_m^{\mathrm{act}},T_m^{\mathrm{life}}$ &
active execution time and total lifetime of container $c_m$ \\

$r_h,r_c$ &
host static-power ratio and container idle-power ratio \\

$E_n^{\mathrm{sta}}$ &
static energy consumption of host $H_n$ \\

$E_m^{\mathrm{act}},E_m^{\mathrm{idle}}$ &
active and idle energy consumption of container $c_m$ \\

$E_n$ &
total energy consumption of host $H_n$ \\

\bottomrule
\end{tabularx}
\end{table}

\subsection{Execution Model}

We first describe the event-driven execution lifecycle and defines the timing quantities used in the model. When workflow $W_k$ arrives at time $A_k$, the scheduler first identifies the tasks that are ready according to the DAG. The TS agent then sequentially assigns these ready tasks to admissible existing containers or requests new containers. All new-container requests generated during the current TS phase are added to a deployment queue. After the TS phase is completed, the CS agent places the corresponding undeployed containers on hosts, and each newly created container becomes operational only after the startup delay $ST_c$.

Because task durations are stochastic, a long predicted queue can accumulate considerable timing error. We therefore restrict each container to at most two unfinished tasks: one task in execution and at most one waiting task. A container whose execution and waiting positions are both occupied is removed from the feasible TS action
set. When the executing task finishes, the waiting task may start only after all required predecessor data have arrived. Otherwise, the container remains idle until the start-time condition in~\eqref{eq_2.2} is satisfied. Workflow $W_k$ is successful only if all of its tasks finish no later than $D_k$.

The timing model distinguishes between quantities available when a scheduling action is chosen and quantities realized later during actual execution. Candidate actions are evaluated from nominal information, whereas the simulator advances according to sampled execution. Candidate actions are evaluated from nominal information, whereas the actual system evolution follows sampled execution capacities. The startup time $ST_c$, execution time $\tau^{\mathrm{ex}}_{k,j}$, transmission time $\tau^{\mathrm{tx}}_{k,ij}$, container-ready time $R_{\mu_{k,j}}$, task start time $S_{k,j}$, task finish time $F_{k,j}$, and workflow finish time $F_k$ together define the event timeline.

\subsubsection{Execution time of tasks}
When task $t_{k,j}$ starts on its assigned container $c_{\mu_{k,j}}$, the task-specific capacity $Q_{k,j}^{(\mu_{k,j})}$ is sampled according to~\eqref{eq:speed}. The realized execution time is then
\begin{equation}
    \tau^{\mathrm{ex}}_{k,j}=\frac{p_{k,j}}{Q_{k,j}^{(\mu_{k,j})}},
    \label{eq_1}
\end{equation}
whereas the scheduler evaluates a candidate container $c_m$ using the predicted execution time $\widehat\tau^{\mathrm{ex}}_{k,j,m}=p_{k,j}/\bar Q_m$. The prediction is available at the decision epoch; the realized capacity is sampled only when execution begins.

\subsubsection{Data transmission time of tasks}
After a predecessor task $t_{k,i}$ finishes, its output data of volume $d_{k,ij}$ must be transmitted to the container hosting its successor task $t_{k,j}$ before $t_{k,j}$ can start. The transmission time depends on the relative placement of the two tasks' containers, and three cases are considered:

\begin{itemize}
\item \textbf{Same container:} If $t_{k,i}$ and $t_{k,j}$ are assigned to the same container, the output data already resides in local memory, so no data transfer is required and the transmission time is zero.

\item \textbf{Different containers on the same host:} If $t_{k,i}$ and $t_{k,j}$ are assigned to different containers co-located on the same host, the data is transferred through the intra-host network (e.g., shared memory bus or virtual bridge) at bandwidth $B^{in}$. The transmission time is $d_{k,ij}/B^{in}$.

\item \textbf{Containers on different hosts:} If the containers hosting $t_{k,i}$ and $t_{k,j}$ are deployed on different physical hosts, the data must traverse the inter-host network link with total bandwidth $B^{cr}$. Because this link is shared among all containers concurrently receiving data on the destination host $H_{\eta_{\mu_{k,j}}}$, each container receives an equal share of the bandwidth. Let $n_{\mathrm{act}}^{\eta_{\mu_{k,j}}}(t)$ denote the number of such receiving containers at transmission start time $t$. The effective bandwidth per container is $B^{cr}/n_{\mathrm{act}}^{\eta_{\mu_{k,j}}}(t)$, yielding a transmission time of
$d_{k,ij}n_{\mathrm{act}}^{\eta_{\mu_{k,j}}}(t)/B^{cr}$.
\end{itemize}

Formally, the transmission time from $t_{k,i}$ to $t_{k,j}$ is
\begin{equation}
    \tau^{\mathrm{tx}}_{k,ij}=
    \begin{cases}
    0, & \mu_{k,i}=\mu_{k,j},\\[2pt]
    \displaystyle\frac{d_{k,ij}}{B^{in}}, & \mu_{k,i}\ne\mu_{k,j},\;
    \eta_{\mu_{k,i}}=\eta_{\mu_{k,j}},\\[2pt]
    \displaystyle\frac{d_{k,ij}\,n_{\mathrm{act}}^{\eta_{\mu_{k,j}}}(t)}{B^{cr}}, & \eta_{\mu_{k,i}}\ne\eta_{\mu_{k,j}}.
    \end{cases}
    \label{eq:tx_time}
\end{equation}

\subsubsection{Ready time of containers}
Only a container with no waiting task can be selected for another assignment. For such a candidate, the ready time is the earliest instant at which the newly assigned task can occupy the execution position:
\begin{equation}
R_{\mu_{k,j}}=
\begin{cases}
T_0+ST_c, & \text{if $c_{\mu_{k,j}}$ is newly deployed},\\
F_{\mu_{k,j}}^{\mathrm{cur}}, & \text{if it is executing one task},\\
t_{\mathrm{now}}, & \text{if it is idle}.
\end{cases}
\label{eq_2.1}
\end{equation}
Here, $T_0$ is the time at which the deployment of $c_{\mu_{k,j}}$ is requested, $F_{\mu_{k,j}}^{\mathrm{cur}}$ is the already scheduled finish time of the task currently executing on $c_{\mu_{k,j}}$, and $t_{\mathrm{now}}$ is the current decision time. The newly assigned task may still start later than $R_{\mu_{k,j}}$ if its predecessor data have not yet arrived.

\subsubsection{Start time of tasks}
When task $t_{k,j}$ is assigned to container $c_{\mu_{k,j}}$, its start time is jointly determined by the container ready time $R_{\mu_{k,j}}$ and the data-ready times of all its predecessors:
\begin{equation}
S_{k,j}=
\begin{cases}
\max\{A_k,R_{\mu_{k,j}}\},
& t_{k,j}=t_{k,\mathrm{entry}},\\[0.5ex]
\max\{
R_{\mu_{k,j}},
\\
\displaystyle\max_{t_{k,i}\in \operatorname{pred}(t_{k,j})}
(F_{k,i}+\tau^{\mathrm{tx}}_{k,ij})
\},
& \text{otherwise}.
\end{cases}
\label{eq_2.2}
\end{equation}
For an entry task, the workflow arrival time $A_k$ acts as its data-ready time, which avoids taking a maximum over an empty predecessor set.

\subsubsection{Finish time of tasks}
The finish time of task $t_{k,j}$ on container $c_{\mu_{k,j}}$ is obtained by adding its realized execution time to its start time:
\begin{equation}
    F_{k,j}=S_{k,j}+\tau^{\mathrm{ex}}_{k,j}.
    \label{eq_2.3}
\end{equation}

\subsubsection{Finish time of workflows}
A workflow is completed when all of its tasks have finished. Therefore, the completion time of workflow $W_k$ is
\begin{equation}
    F_k=\max\limits_{t_{k,j}\in\mathcal T_k}F_{k,j}.
    \label{eq_3}
\end{equation}
Workflow $W_k$ is considered successfully completed if and only if $F_k\leq D_k$.

\subsection{Energy Model}
This subsection formulates the energy consumption model for the cloud service center, where the total energy of a host is decomposed into host static energy, container active energy, and container idle energy. To that end, we first define the power parameters of hosts and containers, then compute each energy component accordingly.

\subsubsection{Power parameters}
The mean power of host $H_n$ under full utilization is denoted by $\bar P_n$. Following the widely adopted linear server power model~\cite{jin2020review}, a host that is powered on but not fully loaded still consumes a static power proportional to $\bar P_n$:
\begin{equation}
    \label{eq:host_static_power}
    P^{\mathrm{sta}}_n=r_h\bar P_n,
\end{equation}
where $r_h\in(0,1)$ is the host static-power ratio, representing the fraction of peak power consumed when the host is idle.

Similarly, the mean power of container $c_m$ is proportional to its share of CPU cores on the hosting host: 
\begin{equation}
    \label{eq:mean_container_power}
    \bar P_m=\frac{\bar P_{\eta_m}(1-r_h)C_m}{C_{\eta_m}}.
\end{equation}
Here, $(1-r_h)$ represents the dynamic portion of the host power that is allocated to containers according to their CPU-core shares.

Under the linear power-performance model based on dynamic voltage and frequency scaling~\cite{9491116,jin2020review}, the dynamic power consumed while task $t_{k,j}$ executes on container $c_m$ scales with the task-specific capacity realization:
\begin{equation}\label{eq:actual_power}
    P_{k,j}^{(m)}=\bar P_m\frac{Q_{k,j}^{(m)}}{\bar Q_m}.
\end{equation}
Consequently, the power consumption can differ between successive tasks on the same container because a new capacity realization is drawn for each execution.

\subsubsection{Host static energy}
A host is considered active while at least one container is deployed on it. During its active period, the host incurs static energy consumption proportional to its powered-on duration:
\begin{equation}
    \label{eq:host_static_energy}
    E_n^{\mathrm{sta}}
    =
    P_n^{\mathrm{sta}}T_n^{\mathrm{on}}
    =
    r_h\bar P_nT_n^{\mathrm{on}},
\end{equation}
where $T_n^{\mathrm{on}}$ denotes the continuous powered-on duration of host $H_n$, from its activation until it is shut down after its last container is terminated.

\subsubsection{Container active energy}
The active energy of container $c_m$ is the sum of the energy consumed by all tasks executed on it:
\begin{equation}
    E_m^{\mathrm{act}}
    =
    \sum_{(k,j):\mu_{k,j}=m}
    P_{k,j}^{(m)}\tau^{\mathrm{ex}}_{k,j}
    =
    \frac{\bar P_m}{\bar Q_m}
    \sum_{(k,j):\mu_{k,j}=m}p_{k,j},
    \label{eq:container_active_energy}
\end{equation}
where the second equality follows from~\eqref{eq_1} and \eqref{eq:actual_power}. Under the adopted linear model, the sampled speed changes power and duration in opposite directions, so their effects cancel for a fixed workload.

\subsubsection{Container idle energy}
Between consecutive task executions, a container remains alive but idle and still consumes a reduced level of power. The idle energy of container $c_m$ is
\begin{equation}
    \label{eq:container_idle_energy}
    E_m^{\mathrm{idle}}
    =
    r_c\bar P_m\left(T_m^{\mathrm{life}}-T_m^{\mathrm{act}}\right),
\end{equation}
where $r_c\in(0,1)$ is the container idle-power ratio, $T_m^{\mathrm{life}}$ is the total lifetime of container $c_m$ from creation to termination, and $T_m^{\mathrm{act}}$ is its cumulative active execution time.

\subsubsection{Total energy consumption}

The total energy consumption of host $H_n$ aggregates the host static energy and the energy consumed by all containers deployed on it:
\begin{equation}
    \label{eq:total_host_energy}
    E_n
    =
    E_n^{\mathrm{sta}}
    +
    \sum_{c_m\in\mathcal C_n^{\mathrm{all}}}
    \left(E_m^{\mathrm{act}}+E_m^{\mathrm{idle}}\right).
\end{equation}

Although $Q_{k,j}^{(m)}$ does not alter the active energy of a fixed workload under the linear model, it changes the execution timeline through $\tau^{\mathrm{ex}}_{k,j}=p_{k,j}/Q_{k,j}^{(\mu_{k,j})}$. The resulting shifts in task completion, data readiness, and container reuse affect container idle durations and host powered-on durations. Therefore, energy differences mainly arise from these timeline effects, the selected container and host types, and the degree of resource reuse, rather than from a direct speed multiplier on active energy.

\subsection{Problem Formulation}
Let $\pi$ denote a scheduling policy that determines both task-to-container assignments and container-to-host placements. Given a workflow set $\mathcal W=\{W_1,\ldots,W_K\}$ arriving over a finite scheduling horizon and a cloud infrastructure $\mathcal H=\{H_1,\ldots,H_N\}$, we aim to find a policy $\pi$ that jointly optimizes three objectives subject to resource-capacity, task-dependency, and assignment constraints.

\subsubsection{Optimization objectives}

The goal is to maximize the workflow scheduling success rate and the container utilization while minimizing the total energy consumption. The three objectives are defined as follows:
\begin{equation}
    \begin{cases}
    J_{\mathrm{suc}}(\pi)=\displaystyle\frac{N_{\mathrm{succ}}}{K}
    \\[1ex]
    J_{\mathrm{uti}}(\pi)=
    \displaystyle
    \frac{1}{|\mathcal C^{\mathrm{all}}|}
    \sum_{c_m\in\mathcal C^{\mathrm{all}}}
    \frac{T_m^{\mathrm{ act}}}{T_m^{\mathrm{life}}}
    \\[1ex]
    J_{\mathrm{ene}}(\pi)=
    \displaystyle\sum_{n=1}^{|\mathcal H|}E_n
    \end{cases}
    \label{eq_5}
\end{equation}
where $N_{\mathrm{succ}}=|\{k:F_k\leq D_k\}|$ is the number of workflows completed before their deadlines. The policy aims to maximize $J_{\mathrm{suc}}(\pi)$ and $J_{\mathrm{uti}}(\pi)$ while minimizing $J_{\mathrm{ene}}(\pi)$.

\subsubsection{Problem constraints}
The scheduling solution must satisfy the following constraints.

\textit{C1 (CPU capacity):} for each host $H_n$ and each time $t$,
\begin{equation}
    \label{eq:C1}
    \sum\limits_{c_m\in\mathcal C_n(t)}C_m\leq C_n
\end{equation}
ensures that the aggregate CPU-core demand of all currently alive containers does not exceed the host capacity.

\textit{C2 (memory capacity):} for each host $H_n$ and each time instance $t$,
\begin{equation}
    \label{eq:C2}
    \sum\limits_{c_m\in\mathcal C_n(t)}M_m\leq M_n
\end{equation}
ensures that the aggregate memory demand of all currently alive containers does not exceed the host capacity.

\textit{C3 (dependency relationship):}
\begin{equation}
    \label{eq:C3}
    S_{k,j}
    \geq
    F_{k,i}+\tau^{\mathrm{tx}}_{k,ij},
    \;
    \forall\,t_{k,i}\in \operatorname{pred}(t_{k,j}),
\end{equation}
requires that the output of every immediate predecessor arrives before task $t_{k,j}$ starts. For an entry task, the predecessor set is empty, so C3 is vacuously satisfied and its start time follows the first branch of~\eqref{eq_2.2}.

\textit{C4 (unique assignment):}
\begin{equation}
    \sum_{m=1}^{|\mathcal C^{\mathrm{all}}|}
    x_{k,j,m}
    \leq 1,
    \label{eq:C4}
\end{equation}
where $x_{k,j,m}\in\{0,1\}$ is a binary assignment variable such that $x_{k,j,m}=1$ if task $t_{k,j}$ is assigned to container $c_m$, and $x_{k,j,m}=0$ otherwise. The inequality allows $\sum_m x_{k,j,m}=0$ when a task is discarded after its workflow deadline, while a scheduled task is assigned to at most one container.

The global objectives in~\eqref{eq_5} can only be evaluated after all workflows have been processed. In practice, however, the policy $\pi$ makes decisions at the granularity of individual tasks and containers: for each ready task, a container is selected; for each new container, a host is selected. The global objectives therefore emerge from the accumulation of these per-step decisions over all scheduling intervals.

Since the scheduling decisions at each interval depend only on the currently observed system state and cannot anticipate future workflow arrivals or container-speed realizations, this problem is a dynamic stochastic optimization that is NP-hard even in its static deterministic variant~\cite{10061217}. Moreover, because scheduling decisions are triggered by workflow-arrival and task-completion events rather than by fixed-duration time slots, the process is modeled as an SMDP. The TS and CS agents have separate decision sequences: TS decisions are generated for the currently ready tasks, whereas CS decisions are generated for undeployed containers produced by new-container requests. After the current TS phase is completed, the corresponding CS placement decisions are processed before physical time advances. Consequently, successive decisions of each agent may be separated by different amounts of physical time.

\section{Scheduling Algorithms}
\label{lab-sec-alg}
The GA-HRL scheduler follows a four-stage procedure that mirrors the scheduling model introduced above. First, the currently active workflows are kept in DAG form and preprocessed to obtain task-level timing indicators from their predicted execution time. Second, a multi-head GAT updates each task representation using information from its dependency neighborhood. Third, the TS agent combines the resulting task representation with the current container state and decides whether a ready task should reuse an admissible existing container or request a new container. Fourth, if a new container is requested, the CS agent selects a feasible host for it. This ordering connects the graph representation directly to the two coupled resource-allocation decisions, rather than treating GAT, hierarchical control, and policy generation as independent components.

The two decision layers operate in the same event-driven environment and can be described as two interacting SMDPs. Workflow arrivals and task completions trigger scheduling events. At each event, the TS agent first processes all currently ready tasks sequentially. New-container requests generated during the TS phase are added to a deployment queue. After the TS phase is completed, the CS agent processes all undeployed containers in the queue and places them on hosts. Physical epochs advances only after both the TS and CS decision phases associated with the current scheduling event have been completed. Because TS and CS are invoked at different frequencies, each agent has its own sequence of decision epochs and its own holding times between successive invocations.

Specifically, the TS Agent operates at its own decision epochs: it observes the dependency‑aware representation of the selected ready task and the admissible containers, then either reuses an existing container with available capacity or requests a new one. Its reward combines predicted deadline margin, uncertainty‑sensitive energy proxies, and container availability. The CS Agent, in contrast, is activated only after the TS phase is completed, and each undeployed container from a new‑container request triggers a CS decision. The CS agent observes container requirements, feasible host capacities, and predecessor locality, then places the container on an existing or newly activated host. If the TS phase produces no new‑container request, no CS decision occurs for that event.

\begin{figure*}[!ht]
    \centering
    \includegraphics[width=0.95\textwidth]{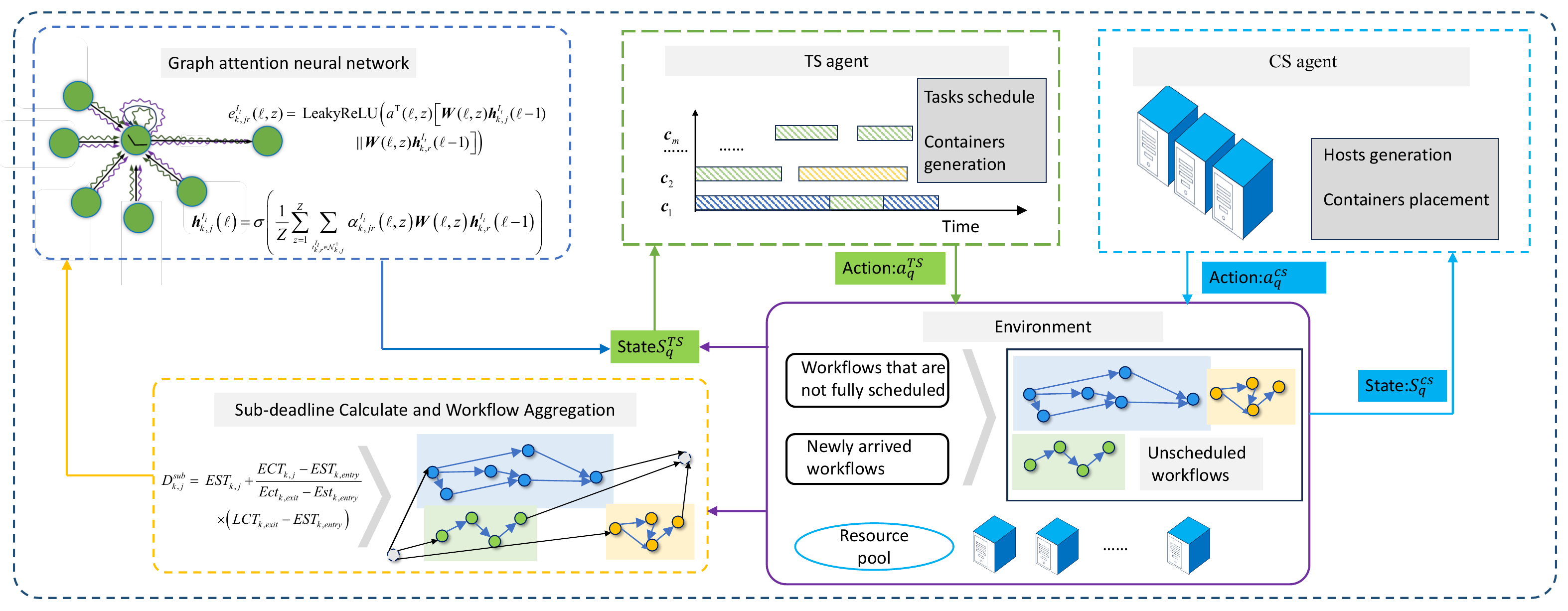} 
    \caption{Schematic diagram of the GA-HRL scheduling algorithm.}
    \label{fig_1}
\end{figure*}

For each agent $g\in\{\mathrm{TS},\mathrm{CS}\}$, the induced process is represented as $\mathcal M^g=(\mathcal S^g,\mathcal A^g,P^g,R^g,\Delta^g)$. Here, $\mathcal S^g$, $\mathcal A^g$, $P^g$, and $R^g$ denote the state space, action space, transition map, and reward function of agent $g$, respectively. Let $T_q^g$ denote the starting time of the $q$-th decision epoch of agent $g$, and let $\Delta_q^g=T_{q+1}^g-T_q^g$ denote the corresponding holding time between two consecutive decisions of the same agent. {The holding time characterizes the irregular physical-time spacing between successive decisions and affects the subsequent state through task execution, data transmission, workflow arrivals, task completions, and resource-state evolution. For policy optimization, the successive invocations of each agent form its decision-epoch trajectory.} Fig.~\ref{fig_1} summarizes this complete path from workflow representation to hierarchical decision making and training.

\subsection{Workflow Embedding for State Construction}
The state representation used by the TS policy is constructed in two complementary steps. The first step computes a reference timing profile for each workflow before task assignment, providing an estimate of task urgency. The second step applies a GAT to the active workflow DAGs, producing task embeddings that combine local task attributes with dependency information. The reference timing profile is not intended to guarantee feasibility under every stochastic execution realization, and the GAT itself does not make resource decisions. Instead, the two parts jointly define the task-level state consumed by the TS policy. 

\subsubsection{Sub-deadline calculation}
Following the timing procedure adopted in the Stochastic Hybrid Workflows Scheduling (SHWS) system~\cite{9491116}, we construct a reference timing profile using execution time that is available before task assignment. Let $\bar Q_{\mathrm{ref}}$ denote the nominal capacity of a high-capacity reference container. The reference execution time of task $t_{k,j}$ is defined as
\begin{equation}
    \widehat\tau^{\mathrm{ex,ref}}_{k,j}=\frac{p_{k,j}}{\bar Q_{\mathrm{ref}}}.
    \label{eq:ref_execution}
\end{equation}
Based on this reference profile, the forward pass computes the Earliest Start Time (EST) of each task:
\begin{align}
& EST_{k,j}= \nonumber \\
& \begin{cases}
& A_k, \quad  \text{if $t_{k,j}$ = $t_{k,\mathrm{entry}}$},\\
& \max\limits_{t_{k,i} \in \operatorname{pred}(t_{k,j})}
\{EST_{k,i}+
\widehat\tau^{\mathrm{ex,ref}}_{k,i}+d_{k,ij}/B^{cr}\}, \text{otherwise}.
\end{cases}
\label{eq_10}
\end{align}
The Earliest Completion Time (ECT) is then
\begin{equation}
    ECT_{k,j} = EST_{k,j}+\widehat\tau^{\mathrm{ex,ref}}_{k,j}.
    \label{eq_11}
\end{equation}

Similarly, the backward pass computes the Latest Completion Time (LCT) of each task:\begin{align}
& LCT_{k,j}= \nonumber \\
& \begin{cases}
& D_k, \quad \text{if $t_{k,j}$ = $t_{k,\mathrm{exit}}$},
\\
& \mathop{\min}\limits_{t_{k,r}\in \operatorname{succ}(t_{k,j})}
\{LCT_{k,r}-
\widehat\tau^{\mathrm{ex,ref}}_{k,r}
-d_{k,jr}/B^{cr}
\},
 \text{otherwise}.
\end{cases}
\label{eq_12}
\end{align}

Using the partial critical path-based deadline-distribution~\cite{8000634} form adopted in SHWS, and taking the workflow entry and exit tasks as global anchors, the sub-deadline of task $t_{k,j}$ is
\begin{equation}
\begin{aligned}
D^{\mathrm{sub}}_{k,j}
=&EST_{k,\mathrm{entry}}
+\frac{ECT_{k,j}-EST_{k,\mathrm{entry}}}
{ECT_{k,\mathrm{exit}}-EST_{k,\mathrm{entry}}}\\
&\times
\left(
LCT_{k,\mathrm{exit}}-EST_{k,\mathrm{entry}}
\right).
\end{aligned}
\label{eq_13}
\end{equation}
These timing quantities are heuristic urgency indicators used in task features and reward shaping, rather than hard guarantees under every stochastic capacity realization. The actual success of a workflow is determined only by the realized completion condition $F_k\leq D_k$.

\subsubsection{Graph embedding with GAT}
The sub-deadline calculation provides each task with an urgency indicator, but it does not yet encode the dependency structure of the workflow. To capture this structural information, we apply a multi-head GAT to the active workflow DAGs following the attention mechanism
in~\cite{velickovic2017graph}.

Because multiple workflows may coexist at a decision epoch, we merge all active workflows into a single DAG. Let $I_t$ denote the current event-driven decision epoch. At epoch $I_t$, a pseudo-entry task $t_{\mathrm{psd\text{-}enter}}$ is prepended to all entry tasks, and a
pseudo-exit task $t_{\mathrm{psd\text{-}exit}}$ is appended after all exit tasks. Both pseudo tasks have zero execution and transfer cost, and their initial feature vectors are set to zero because they do not correspond to real computational tasks. These auxiliary nodes are used only to connect multiple active workflows into one graph and are never scheduled as real tasks. Through the above processing, we can use GAT on multiple workflows, and the
detailed information is shown in Fig.~\ref{fig_1}.

For each original task $t_{k,j}$ in the merged graph, we denote its node by
$t^{I_t}_{k,j}$. To characterize the communication demand from its
immediate predecessors, we define the maximum predecessor data volume as
\begin{equation}
d_{k,j}^{\mathrm{pred,max}}=
\begin{cases}
\displaystyle
\max_{t_{k,i}\in \operatorname{pred}(t_{k,j})} d_{k,ij},
& \operatorname{pred}(t_{k,j})\neq\emptyset,\\[1ex]
0, & \text{otherwise}.
\end{cases}
\label{eq:pred_data_proxy}
\end{equation}

The initial feature vector of task $t^{I_t}_{k,j}$ is then
\begin{equation}
\bm h^{I_t}_{k,j}(0)=
\left[
D^{\mathrm{sub}}_{k,j},
EST_{k,j},
d_{k,j}^{\mathrm{pred,max}},
p_{k,j},
\left|\operatorname{succ}(t^{I_t}_{k,j})\right|
\right],
\label{eq_14}
\end{equation}
where $D^{\mathrm{sub}}_{k,j}$ is the sub-deadline obtained by assuming
that $t^{I_t}_{k,j}$ is scheduled to a newly created container with the
best computational capacity, and
$|\operatorname{succ}(t^{I_t}_{k,j})|$ is the number of immediate
successors of $t^{I_t}_{k,j}$.

The GAT encoder consists of several stacked graph attention layers. Let $N^{I_t}=|\mathcal T^{I_t}|$ denote the number of nodes in the merged DAG at epoch $I_t$. The first-layer input feature is
\begin{equation}
    \bm{H}^{\mathrm{gat}}(0) =
    \left[ \bm{h}^{I_t}_1(0), \bm{h}^{I_t}_2(0), \dots, \bm{h}^{I_t}_{N^{I_t}}(0) \right]^\top,
    \label{eq_15}
\end{equation}
where $d_0$ is the initial feature dimension. At layer $\ell$, the GAT layer updates each node representation by attending over its neighbors, producing the refined feature matrix
\begin{equation}
    \bm H^{\mathrm{gat}}(\ell)
    =
    \left[
    \bm h^{I_t}_1(\ell),\;
    \ldots,\;
    \bm h^{I_t}_{N^{I_t}}(\ell)
    \right]^\top
    \in \mathbb{R}^{N^{I_t}\times d_h}.
    \label{eq_14.1}
\end{equation}
where $d_h$ denotes the embedding dimension at the output of layer $\ell$.

To retain the node's own information during aggregation and to avoid an empty neighborhood for exit nodes, we define the augmented successor neighborhood
\begin{equation}
    \mathcal N_{k,j}^{+}=
    \operatorname{succ}(t^{I_t}_{k,j})\cup\left\{t^{I_t}_{k,j}\right\}.
\end{equation}
For head $z$ of layer $\ell$, the attention coefficient between
$t^{I_t}_{k,j}$ and a node $t^{I_t}_{k,r}\in\mathcal N_{k,j}^{+}$ is
\begin{equation}
    \alpha^{I_t}_{k,jr}(\ell,z)=
    \frac{
    \exp\left(e^{I_t}_{k,jr}(\ell,z)\right)
    }{
    \displaystyle
    \sum\nolimits_{t^{I_t}_{k,r'}\in\mathcal N_{k,j}^{+}}
    \exp\left(e^{I_t}_{k,jr'}(\ell,z)\right)
    }.
    \label{eq_14.2}
\end{equation}

The scalar attention score is computed as
\begin{align}
    e^{I_t}_{k,jr}(\ell,z)
    &=
    \operatorname{LeakyReLU}
    \Big(
    \bm a^{\top}(\ell,z)
    \big[
    \bm W(\ell,z)\bm h^{I_t}_{k,j}(\ell-1) \nonumber\\
    &\qquad\mathbin{\Vert}
    \bm W(\ell,z)\bm h^{I_t}_{k,r}(\ell-1)
    \big]
    \Big),
    \label{eq_14.3}
\end{align}
where $\bm W(\ell,z)$ is the learnable feature-transformation matrix and $\bm a(\ell,z)$ is the learnable attention vector of head $z$ in layer $\ell$. The node representation is then updated by multi-head aggregation over $\mathcal N_{k,j}^{+}$:
\begin{align}
    \bm h^{I_t}_{k,j}(\ell)
    =
    \sigma
    \Bigg(
    \frac{1}{Z}\sum_{z=1}^{Z}
    \sum_{t^{I_t}_{k,r}\in\mathcal N_{k,j}^{+}}
    \alpha^{I_t}_{k,jr}(\ell,z)
    \bm W(\ell,z)
    \bm h^{I_t}_{k,r}(\ell-1)
    \Bigg),
    \label{eq_14.4}
\end{align}
where $Z$ is the number of attention heads and $\sigma(\cdot)$ is the activation function.

We denote the resulting stacked GAT encoder by $f_\psi^{\mathrm{GAT}}$, where $\psi$ collects all learnable parameters $\{\bm W(\ell,z),\bm a(\ell,z)\}_{\ell,z}$.

\subsection{Task Scheduling (TS) Agent}
The TS agent determines the container assignment for each ready task according to both the task characteristics and the current container state. This subsection defines its state representation, action space, and reward function.

\subsubsection{State representation}
At a TS decision epoch, the TS agent sequentially processes all
currently ready tasks before physical time advances. For task
$t_{k,j}$ scheduled at step $q$, the state $s^{\mathrm{TS}}_q$ contains the following information:
\begin{itemize}
    \item the embedding vector of the selected task $t_{k,j}$;
    \item for each existing candidate container $c_m$, its hosting host $\eta_m$, CPU-core allocation $C_m$, memory allocation $M_m$, mean computational capacity $\bar Q_m$, execution-speed variation coefficient $v$, current ready time or workload, and the estimated predecessor-to-container data-transfer delays under the current placement;
    \item for each candidate new-container type, its CPU-core allocation, memory allocation, nominal computational capacity, and execution-speed variation coefficient $v$. The host-dependent transmission information of a new container is determined only after the CS agent selects its host.
\end{itemize}

\subsubsection{Action space}
At TS decision step $q$, the agent chooses either to assign the
selected task to an admissible existing container or to request a new container of a specified type. Let $\mathcal C_q^{\mathrm{feas}}$ denote the set of existing containers that are feasible under the current scheduling rules, $N_c^{\mathrm{type}}$ the number of available container types, and $\iota_i$ the action of creating a new container of type $i$. The TS action space ($a_q^{\mathrm{TS}}\in\mathcal A_q^{\mathrm{TS}}$) is 
\begin{equation}
    \mathcal A_q^{\mathrm{TS}}
    =
    \mathcal C_q^{\mathrm{feas}}
    \cup
    \left\{\iota_i\mid i=1,\ldots,N_c^{\mathrm{type}}\right\}.
    \label{eq_16.0}
\end{equation}

An existing container belongs to $\mathcal C_q^{\mathrm{feas}}$ only if it has no waiting task; hence it contains either one executing task or is idle. A container that already has one executing task and one waiting task is masked out before policy sampling. If $a_q^{\mathrm{TS}}=c_m$, the selected task occupies the sole waiting position when $c_m$ is busy, or the execution position when $c_m$ is idle. If $a_q^{\mathrm{TS}}=\iota_i$, a new container of type $i$ is
requested and added to the deployment queue. After the current TS phase is completed, the corresponding undeployed container is subsequently placed by the CS agent.

\subsubsection{Reward function}
For task $t_{k,j}$ and candidate container $c_m$, the TS reward
evaluates the expected scheduling quality of each feasible action:
\begin{align}
&r^{\mathrm{TS}}_q
=w_1^{\mathrm{TS}}
\frac{\left(D^{\mathrm{sub}}_{k,j}-EST_{k,j}\right)(1+v)}
{\widehat\tau^{^{\mathrm{ex}}}_{k,j,m}}
\nonumber\\
&\quad+w_2^{\mathrm{TS}}
\left(
1-
\frac{\bar E_m^{\mathrm{act}}+\bar E_m^{\mathrm{idle}}}
{\bar E_{\mathrm{act,sub}}+\bar E_{\mathrm{idle,sub}}}
\right)
+w_3^{\mathrm{TS}}
\frac{EST_{k,j}-R_m}{T_{\mathrm{norm}}},
\label{eq_16}
\end{align}
where $\widehat\tau^{\mathrm{ex}}_{k,j,m}=p_{k,j}/\bar Q_m$ is the predicted execution time. The realized time
$p_{k,j}/Q_{k,j}^{(m)}$ is sampled only when the task actually starts and is not used to rank actions. The container-availability term is positive when $c_m$ is expected to be ready before $EST_{k,j}$ and becomes negative when the task would wait beyond that reference time.

For an existing container, $\bar Q_m$ and $\bar P_m$ are evaluated using its actual hosting host. For a newly requested container whose placement has not yet been determined, the TS agent evaluates the reward by assuming that the container is deployed on the feasible host that yields the minimum estimated energy consumption. This assumption is used only to evaluate the new-container action; the actual host is subsequently selected by the CS agent.

The factor $(1+v)$ is introduced as an uncertainty-sensitive term for reward shaping. It should not be interpreted as a probabilistic upper bound or as a physical correction applied to the execution-time or energy model. Its purpose is to give more weight to preserving sub-deadline margin when the execution-speed variation becomes larger. Similarly, $\bar E_m^{\mathrm{act}}$ and $\bar E_m^{\mathrm{idle}}$ are
energy proxies used to rank candidate actions rather than realized energy values. We define
\begin{equation}
    \bar E_m^{\mathrm{act}}
    =
    \frac{\bar P_m p_{k,j}(1+v)}{\bar Q_m},
    \label{eq:ts_active_proxy}
\end{equation}
solely to introduce an uncertainty-sensitive penalty into the reward. This proxy does not contradict the system energy model, in which the realized active energy of a fixed workload is independent of the sampled speed under the linear power-performance assumption.

Using the maximum predecessor data volume
$d_{k,j}^{\mathrm{pred,max}}$ defined in~\eqref{eq:pred_data_proxy},
we set
$\bar E_m^{\mathrm{idle}}
=\bar P_m r_c d_{k,j}^{\mathrm{pred,max}}/B^{cr}$. The corresponding values for the fastest reference container are $\bar E_{\mathrm{act,sub}}$ and $\bar E_{\mathrm{idle,sub}}$. The coefficients $w_1^{\mathrm{TS}}$, $w_2^{\mathrm{TS}}$, and $w_3^{\mathrm{TS}}$ control the three terms, and $T_{\mathrm{norm}}$ is a constant used to normalize the container-availability term.

\subsection{Container Scheduling (CS) Agent}

When the TS agent requests a new container, the request is added to the deployment queue. After the current TS phase is completed, the CS agent selects an appropriate host for each undeployed container according to the container requirements and the current host states. If no existing host satisfies the deployment requirements, the CS agent activates a new host. This subsection defines the CS state representation, action space, and reward function.

\subsubsection{State representation}
After the TS agent completes the current ready-task assignment phase, each undeployed container waiting for host placement triggers one CS decision. Thus, a TS phase may be followed by zero, one, or multiple CS decisions, depending on the number of new-container requests generated during task scheduling. The CS agent therefore maintains its own event-driven decision sequence.

For container $c_m$ scheduled at step $q$, the state $s^{\mathrm{CS}}_q$ includes the following information:
\begin{itemize}
    \item the resource requirements of the selected container, including its CPU-core requirement, memory requirement, nominal computational capacity, and the data-transfer speeds achievable on different hosts;
    \item the current workload of each host, including the number of occupied CPU cores, occupied memory, available CPU percentage, available memory percentage, CPU utilization, and mean energy consumption;
    \item the host and container IDs of the critical predecessor task. For each predecessor $t_{k,i}\in \operatorname{pred}(t_{k,j})$, its data-ready time for the current task is estimated from its start time, the execution time $p_{k,i}/\bar Q_{\mu_{k,i}}$ on its assigned container, and the corresponding data-transfer time to $t_{k,j}$. The predecessor with the largest estimated data-ready time is regarded as the critical predecessor, because the current task cannot start until the outputs of all its predecessors have arrived.
\end{itemize}

\subsubsection{Action space}
Let $C_n^{\mathrm{used}}(q)$ and $M_n^{\mathrm{used}}(q)$ denote the number of occupied CPU cores and the occupied memory on host $H_n$ at CS decision step $q$. The feasible existing-host set for container $c_m$ is
\begin{align}
    & \mathcal H_q^{\mathrm{feas}}
    = \nonumber \\
    & `\big\{H_n\in\mathcal H:
    C_n-C_n^{\mathrm{used}}(q)\ge C_m,
    M_n-M_n^{\mathrm{used}}(q)\ge M_m
    \big\}.
    \label{eq:cs_feasible_hosts}
\end{align}

Let $N_h^{\mathrm{type}}$ denote the number of available host types, and let $\xi_i$ denote the action of activating a new host of type $i$. The CS action space is
\begin{equation}
\mathcal A_q^{\mathrm{CS}}
=
\mathcal H_q^{\mathrm{feas}}
\cup
\left\{\xi_i\mid i=1,\ldots,N_h^{\mathrm{type}}\right\},
a_q^{\mathrm{CS}}\in\mathcal A_q^{\mathrm{CS}}.
\label{eq_17}
\end{equation}

Before policy sampling, an action mask removes all existing hosts that violate either the CPU-core or memory requirement of the selected container. Therefore, infeasible host-placement actions cannot be selected by the CS policy.

\subsubsection{Reward function}
Among feasible host-placement actions, the CS agent aims to improve data locality and resource consolidation. Let $H^{\mathrm{crit}}_{k,j}$ denote the host on which the critical predecessor of task $t_{k,j}$ is deployed, and define the locality indicator
\begin{equation}
    x_q^{\mathrm{loc}}=
    \begin{cases}
    1, & \text{if container $c_m$ is deployed on $H^{\mathrm{crit}}_{k,j}$},\\
    0, & \text{otherwise}.
    \end{cases}
    \label{eq_19}
\end{equation}
For an entry task with no predecessor, no critical predecessor exists, and $x_q^{\mathrm{loc}}$ is set to zero.

The CS reward is then
\begin{equation}
    r^{\\mathrm{CS}}_q=
    w_1^{\mathrm{CS}}x_q^{\mathrm{loc}}
    +w_2^{\mathrm{CS}}\frac{C^{\mathrm{used}}_{\eta_m}(q)+C_m}{C_{\eta_m}},
    \label{eq_18}
\end{equation}
where $w_1^{\mathrm{CS}}$, and $w_2^{\mathrm{CS}}$ are the reward coefficients of the CS agent. $C_{\eta_m}$ and $C^{\mathrm{used}}_{\eta_m}(q)$ are the total and used CPU cores of the host selected by the action, respectively, and $C_m$ is the CPU-core requirement of container $c_m$. The first term rewards data locality with the critical predecessor, while the second term promotes resource consolidation among feasible hosts.

The complete online scheduling procedure of GA-HRL is summarized in Algorithm~\ref{alg:gahrl_online}. The algorithm combines the workflow representation, TS decisions, CS decisions, and the subsequent timing and energy updates into one event-driven procedure.

\begin{algorithm}[!t]
\caption{GA-HRL Online Scheduling.}
\label{alg:gahrl_online}
\begin{algorithmic}[1]

\REQUIRE Trained TS policy $\pi_{\theta_{\mathrm{TS}}}$, trained CS policy $\pi_{\theta_{\mathrm{CS}}}$, and trained GAT encoder $f_\psi^{\mathrm{GAT}}$;
workflow stream $\mathcal W$; resource pool $\mathcal H$.
\ENSURE  Task-to-container assignments and container-to-host placements.

\STATE Initialize the scheduling environment and resource states;

\WHILE{the scheduling episode is not terminated}

    \STATE Process workflow arrivals and task completions; for newly
    arrived workflows, compute timing indicators using
    (\ref{eq_10})-(\ref{eq_13}) and update ready set $\mathcal R$;

    \STATE Construct the merged DAG and update task embeddings using the trained GAT encoder $f_\psi^{\mathrm{GAT}}$ according to (\ref{eq_14})-(\ref{eq_14.4});

    \WHILE{$\mathcal R\neq\emptyset$}
        \STATE Select $t_{k,j}$, construct $s_q^{\mathrm{TS}}$
        from (\ref{eq_16.0}), and sample
        $a_q^{\mathrm{TS}}\sim
        \pi_{\theta_{\mathrm{TS}}}(\cdot|s_q^{\mathrm{TS}})$;

        \STATE Execute $a_q^{\mathrm{TS}}$ by reusing a feasible container
        or requesting a new one; enqueue the new container if requested;
    \ENDWHILE

    \WHILE{the deployment queue is not empty}
        \STATE Select $c_m$; construct $s_q^{\mathrm{CS}}$ using
        (\ref{eq:cs_feasible_hosts})-(\ref{eq_17}), and sample
        $a_q^{\mathrm{CS}}\sim
        \pi_{\theta_{\mathrm{CS}}}(\cdot|s_q^{\mathrm{CS}})$;

        \STATE Deploy $c_m$ to the selected feasible host or activate
        a new host;
    \ENDWHILE

    \STATE Advance to the next scheduling event; update execution,  transmission, and task timing using (\ref{eq:speed}), (\ref{eq_1}), (\ref{eq:tx_time}), and (\ref{eq_2.1})-(\ref{eq_3}), and update energy using
    (\ref{eq:host_static_energy})-(\ref{eq:total_host_energy});
\ENDWHILE

\STATE \textbf{return} the final scheduling decisions;

\end{algorithmic}
\end{algorithm}

\subsection{Policy Derivation with PPO}
Both the TS and CS agents are trained with PPO~\cite{schulman2017ppo}. Although the underlying scheduling environment is event-driven and successive decision epochs may be separated by nonuniform physical holding times, each invocation of an agent is treated as one step in that agent's decision-epoch trajectory. Accordingly, we optimize the decision-indexed discounted return:
\begin{equation}
    J^g(\pi_g)
    =
    \mathbb E_{\pi_g}
    \left[
    \sum_{q=0}^{Q_g-1}\gamma^q r_q^g
    \right],
    \label{eq:decision_return}
\end{equation}
where $Q_g$ is the number of decisions made by agent $g$ in an episode, and $\gamma\in(0,1)$ discounts successive decisions rather than elapsed physical time. The holding time $\Delta_q^g$ affects the transition to the next decision state through the physical system evolution, but is not used as an additional time-dependent discount. 

\subsubsection{TD residual and GAE}
In the following, $g\in\{\mathrm{TS},\mathrm{CS}\}$ denotes the agent, and $q$ indexes its consecutive decisions. The Temporal-Difference (TD) residual of agent $g$ is
\begin{equation}
    \delta_q^g
    =r_q^g + m_q^g \gamma V_{\phi_g}(s_{q+1}^g) - V_{\phi_g}(s_q^g),
    \label{eq:smdp_td}
\end{equation}
where $r_q^g$ is the reward after decision $q$, $V_{\phi_g}(\cdot)$ is the critic, and $m_q^g$ is a non-terminal mask with $m_q^g=0$ for terminal transitions and $m_q^g=1$ otherwise. $\gamma\in(0,1)$ is the per-decision discount factor. The generalized advantage estimate (GAE) is then
\begin{equation}
    \hat{A}_q^g
    =
    \delta_q^g
    +
    m_q^g\gamma\lambda_{\mathrm{GAE}}\hat{A}_{q+1}^g,
    \label{eq:smdp_gae}
\end{equation}
where $\lambda_{\mathrm{GAE}}$ controls the bias-variance tradeoff.

\subsubsection{PPO objective}
For both agents, the actor maximizes the clipped surrogate objective
\begin{equation}
    L_{\pi}^{g}(\theta_g)=\hat{\mathbb{E}}\Big[
    \min\big(u_q^g(\theta_g)\hat{A}_q^g,
    \mathrm{clip}(u_q^g(\theta_g),1-\varepsilon,1+\varepsilon)
    \hat{A}_q^g
    \big)\Big],
    \label{eq_20}
\end{equation}
where
\begin{equation}
    u_q^g(\theta_g) = \frac{\pi_{\theta_g}(a_q^g|s_q^g)}
    {\pi_{\theta_g^{\mathrm{old}}}(a_q^g|s_q^g)}
    \label{eq_21}
\end{equation}
is the probability ratio between the current and previous policies, and $\mathrm{clip}(\cdot)$ constrains this ratio to $[1-\varepsilon,1+\varepsilon]$.

\subsubsection{Critic and combined objective}
The target used to train the critic is
\begin{equation}
    \hat{G}_q^g
    =
    \hat{A}_q^g+V_{\phi_g}(s_q^g),
    \label{eq:smdp_target}
\end{equation}
and the critic minimizes the mean-squared value loss
\begin{equation}
L_V^g(\phi_g)
=
\hat{\mathbb E}
\left[
\left(
V_{\phi_g}(s_q^g)-\hat G_q^g
\right)^2
\right].
\label{eq:smdp_value_loss}
\end{equation}

The actor-critic parameters are updated by maximizing the combined PPO objective:
\begin{align}
    J_{\mathrm{PPO}}^g(\theta_g,\phi_g)
    =
    L_\pi^g(\theta_g)
    -c_1L_V^g(\phi_g)
    +c_2\widehat{\mathbb E}
    \left[
    \operatorname{Ent}\!\left(\pi_{\theta_g}(\cdot\mid s_q^g)\right)
    \right],
    \label{eq:ppo_total}
\end{align}
where $c_1$ is the value-loss coefficient, $\operatorname{Ent}(\cdot)$ is policy entropy, and $c_2$ (set to 0.01) is the entropy coefficient used in the implementation.

\subsubsection{Alternating training}
The two agents are trained alternately using separate actor-critic networks and rollout buffers. The alternating PPO training procedure is summarized in Algorithm~\ref{alg:ts_cs_ppo}.

\begin{algorithm}[!t]
\caption{Alternating PPO Training of TS and CS Agents.}
\label{alg:ts_cs_ppo}
\begin{algorithmic}[1]

\REQUIRE GAT encoder $f_\psi^{\mathrm{GAT}}$;
TS policy-value pair $(\pi_{\theta_{\mathrm{TS}}},V_{\phi_{\mathrm{TS}}})$;
CS policy-value pair $(\pi_{\theta_{\mathrm{CS}}},V_{\phi_{\mathrm{CS}}})$;
$\gamma$, $\lambda_{\mathrm{GAE}}$, $\varepsilon$;
number of alternating epochs $E$;
rollout lengths $L_{\mathrm{TS}}$ and $L_{\mathrm{CS}}$;
PPO optimization epochs $S$.

\ENSURE Trained GAT encoder $f_\psi^{\mathrm{GAT}}$,
TS policy $\pi_{\theta_{\mathrm{TS}}}$, and CS policy $\pi_{\theta_{\mathrm{CS}}}$.

\STATE Initialize the GAT encoder, the two actor-critic networks,
and rollout buffers $\mathcal D_{\mathrm{TS}}$ and $\mathcal D_{\mathrm{CS}}$;

\FOR{$e=1$ to $E$}

    \STATE Fix $\pi_{\theta_{\mathrm{CS}}}$ and reset the environment;

    \FOR{$q=1$ to $L_{\mathrm{TS}}$}
        \STATE Compute the current task embedding with         $f_{\psi^{\mathrm{old}}}^{\mathrm{GAT}}$ and construct         $s_q^{\mathrm{TS}}$;
        \STATE Sample $a_q^{\mathrm{TS}}$ from         $\pi_{\theta_{\mathrm{TS}}^{\mathrm{old}}}
        (\cdot|s_q^{\mathrm{TS}})$ and execute the action;
        \STATE Use fixed $\pi_{\theta_{\mathrm{CS}}}$ for intervening CS decisions;
        \STATE Store the TS transition in $\mathcal D_{\mathrm{TS}}$;
    \ENDFOR

    \STATE Calculate $\hat A_q^{\mathrm{TS}}$ and $\hat G_q^{\mathrm{TS}}$;

    \FOR{$i=1$ to $S$}
        \STATE Jointly update $\psi$, $V_{\phi_{\mathrm{TS}}}$, and    $\pi_{\theta_{\mathrm{TS}}}$ using the TS PPO objective;
    \ENDFOR

    \STATE Fix the updated $f_\psi^{\mathrm{GAT}}$ and TS actor-critic network, and reset the environment;

    \FOR{$q=1$ to $L_{\mathrm{CS}}$}
        \STATE Use the fixed TS policy and encoder until a CS decision is triggered;
        \STATE Construct $s_q^{\mathrm{CS}}$;
        \STATE Sample $a_q^{\mathrm{CS}}$ from
        $\pi_{\theta_{\mathrm{CS}}^{\mathrm{old}}}
        (\cdot|s_q^{\mathrm{CS}})$ and execute the action;
        \STATE Store the CS transition in $\mathcal {D}_{\mathrm{CS}}$;
    \ENDFOR

    \STATE Calculate $\hat A_q^{\mathrm{CS}}$ and $\hat G_q^{\mathrm{CS}}$;

    \FOR{$i=1$ to $S$}
        \STATE Update $V_{\phi_{\mathrm{CS}}}$ and
        $\pi_{\theta_{\mathrm{CS}}}$ using the CS PPO objective;
    \ENDFOR

    \STATE Update policy/encoder parameters, clear both buffers,
    and evaluate the joint policy;

\ENDFOR

\end{algorithmic}
\end{algorithm}

In implementation, scheduling one task or deploying one container constitutes one decision step in the corresponding agent-specific trajectory. During the TS training phase, the current CS policy is kept fixed and is used to process the intervening CS decisions required to advance the shared environment. After the TS rollout is collected, the TS agent computes its GAE advantages and updates its actor-critic network using the TS rollout buffer. The updated TS policy is then kept fixed during the subsequent CS training phase, in which it processes the intervening task-scheduling decisions. After the CS rollout is collected, the CS agent computes its GAE advantages and updates its actor-critic network using the CS rollout buffer.

For the finite-horizon scheduling episodes considered here, the number of workflows, tasks, and corresponding TS/CS decisions is finite. With bounded rewards and $0<\gamma<1$, the decision-indexed return in~\eqref{eq:decision_return} is well defined. For a fixed policy, standard policy evaluation under the per-decision discount retains the usual contraction property, so the nonuniform physical holding times do not alter the policy-evaluation formulation adopted here. Since PPO uses nonlinear function approximation and the two policies are coupled through the scheduling environment, however, we do not claim global convergence to an optimal joint policy. Instead, the training curves in our experiments (e.g., Fig.~\ref{Fig_3}) are used to evaluate empirical convergence.

\section{Performance Evaluation}
\label{lab-sec-experiment}
We evaluate GA-HRL using three metrics: workflow scheduling success rate, average container resource utilization, and total system energy consumption.

\subsection{Experimental Setup}
\subsubsection{Environment Configuration}
All experiments were implemented in Python 3.11.13 with \texttt{PyTorch} 2.6.1 and \texttt{sb3-contrib} 2.7. Each reported metric is averaged over 50 independent evaluation runs with different random seeds, covering random workflow arrivals, container execution-speed variations, and learning-based scheduling decisions.

At the beginning of each scheduling episode, no host is active. Hosts are activated on demand by the CS agent. A container is terminated when it has neither an executing task nor a waiting task, and a host is shut down when no container remains on it. If the same physical resource is activated again later, it is treated as a new host instance with a new host ID for scheduling and energy accounting. Accordingly, $N$ denotes the total number of host instances activated during an episode.

\subsubsection{Parameter Settings}
We construct a trace-driven environment from the 2018 Alibaba cluster trace~\cite{alibaba}. Eight container types are considered, with CPU allocations $\{1,2,4,6,8,16,24,32\}$ cores and memory allocations $\{4,8,16,24,32,64,96,128\}$~GB. The mean capacity and power of each
container are derived from its hosting host and CPU-core share. The host configurations are summarized in Table~\ref{tab2}. Following~\cite{10509784}, we use 5,200 workflow DAGs from the trace, each containing at least 10 tasks. The output data volume of each task is uniformly distributed over $[500,5000]$~MB, the cross-host bandwidth is 200~MB/s, and the intra-host bandwidth is 500~MB/s.
\begin{table}[!t]
\centering
\caption{Parameters of host types}
\label{tab2}
\scriptsize
\setlength{\tabcolsep}{4pt}

\begin{tabular}{@{}ccccc@{}}
\toprule
Type & CPU cores &
\makecell[c]{Mean capacity\\(MIPS)} &
Memory (GB) &
\makecell[c]{Mean power\\(W)} \\
\midrule
1 & 96  & 264000 & 384 & 795  \\
2 & 192 & 528000 & 768 & 1600 \\
\bottomrule
\end{tabular}
\end{table}

For each task execution, the capacity $Q_{k,j}^{(m)}$ is sampled from the rejection-sampled model in~\eqref{eq:speed}. The execution-speed variation coefficient is $v\in\{0,\ldots,0.45\}$ with a step size of 0.05. The number of workflows is $K\in\{100,\ldots,1000\}$ with a step size of 100, and workflow arrivals follow a Poisson process with rate $\lambda=0.5$.

The TS and CS agents use the same PPO hyperparameters: learning rate $1\times10^{-5}$, discount factor $\gamma=0.99$, GAE parameter $\lambda_{\mathrm{GAE}}=0.95$, clipping range $\varepsilon=0.2$, value-loss coefficient $c_1=0.5$, entropy coefficient $c_2=0.01$, and
maximum gradient norm $0.5$. Each PPO rollout contains 2,048 decision steps, with a minibatch size of 32 and two optimization epochs per update. The reward coefficients are
$w_1^{\mathrm{TS}}=0.4$,
$w_2^{\mathrm{TS}}=0.3$,
$w_3^{\mathrm{TS}}=0.3$, $w_1^{\mathrm{CS}}=0.7$, and $w_2^{\mathrm{CS}}=0.3$. Training is performed for 1,000 alternating epochs, with 2,048 TS and 2,048 CS training steps per epoch.

Following~\cite{9491116}, the deadline of workflow $W_k$ is 
\begin{equation}
    D_k=A_k+\alpha^{df}S^{\mathrm{fast}}_k,
    \label{eq_47}
\end{equation}
where $\alpha^{df}\in\{2.0,\ldots,2.9\}$ (with a step size of 0.1) is the deadline factor, and $S_k^{\mathrm{fast}}$ is the critical-path length under the fastest-resource assumption. Let $\mathcal P_k$ denote the set of entry-to-exit paths of $W_k$, and let $\mathcal T(\mathcal P)$ and $\mathcal E(\mathcal P)$ denote the task and edge sets on path
$\mathcal P$. Then
\begin{equation}
    S_k^{\mathrm{fast}}
    =
    \max_{\mathcal P\in\mathcal P_k}
    \left(
    \sum_{t_{k,i}\in\mathcal T(\mathcal P)}\tau^{\mathrm{ ex,min}}_{k,i}
    +
    \sum_{e_{k,ij}\in\mathcal E(\mathcal P)}\frac{d_{k,ij}}{B^{cr}}
    \right),
    \label{eq_48}
\end{equation}
where $\tau^{\mathrm{ex,min}}_{k,i}$ is the execution time of $t_{k,i}$ on the fastest container type. Thus, a larger $\alpha^{df}$ relaxes the deadline.

\subsection{Ablation Study}
\begin{figure}[!ht]
\centering
\includegraphics[width=3.4 in]{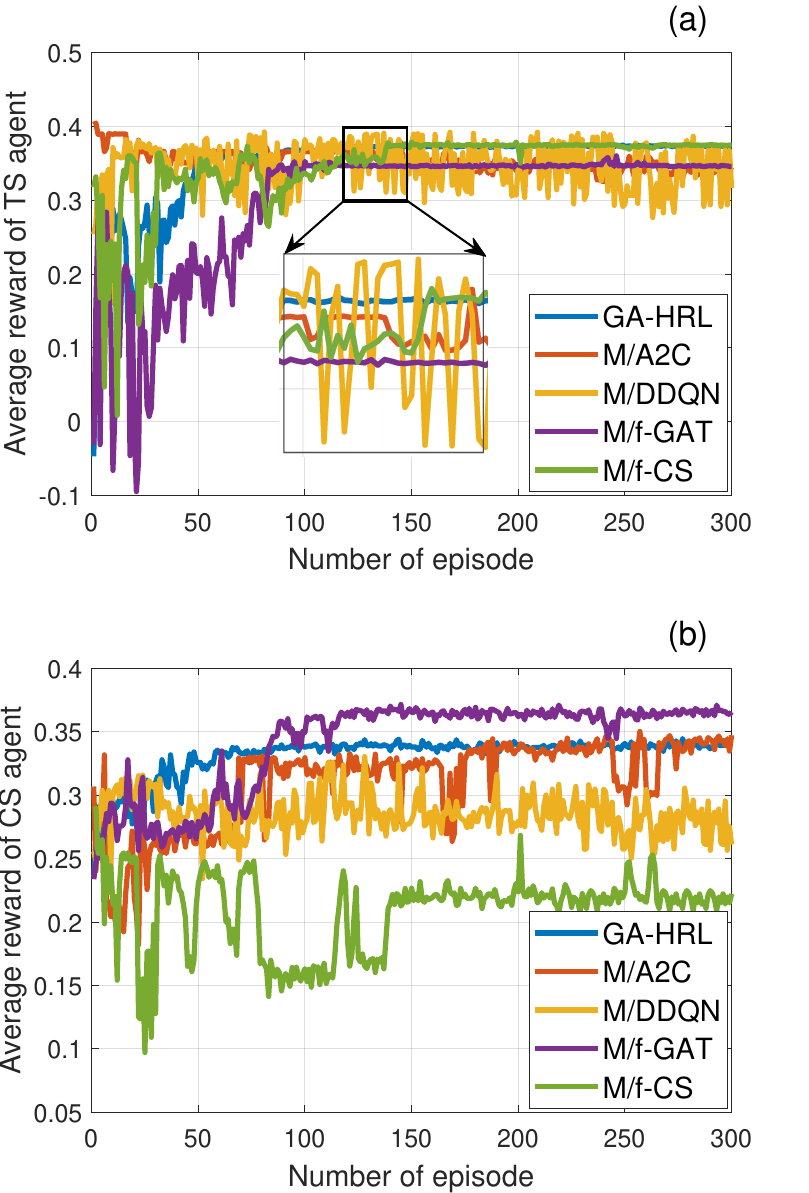}
\caption{Training convergence of different methods: (a) average reward
of the TS agent and (b) average reward of the CS agent.}
\label{Fig_3}
\end{figure}

We construct four ablated versions to examine the contribution of each main component. M/A2C and M/DDQN replace PPO with A2C and DDQN, respectively. M/f-GAT removes the GAT encoder and uses the original task features directly as policy input. M/f-CS retains the
learned TS policy but replaces the CS policy with a random feasible host-placement rule. Fig.~\ref{Fig_3} compares training rewards, and Table~\ref{tab3} reports the final scheduling metrics.

\begin{table}[!htb]
  \centering
  \caption{Ablation study results of GA-HRL}
  \label{tab3}
\begin{tabular}{lccc}
  \toprule
Method & \makecell[c]{Workflow \\ success (\%)} &
\makecell[c]{Container resource \\ utilization (\%)} &
\makecell[c]{Energy \\ consumption (J)} \\
  \midrule
GA-HRL     &100 &  65& \makecell[c]{1.84$\times {10^{7}} $}\\
M/A2C     &97&  64& \makecell[c]{1.98$\times {10^{7}} $}\\
M/DDQN     &95 &  66& \makecell[c]{2.02$\times {10^{7}} $}\\
M/f-GAT     &100 &  65&\makecell[c]{2.03$\times {10^{7}} $}\\
M/f-CS     &92 &  70& \makecell[c]{2.19$\times {10^{7}} $}\\
  \bottomrule
  \end{tabular}
\end{table}

The training curves in Fig.~\ref{Fig_3} show that GA-HRL and M/A2C converge to relatively high TS rewards, whereas M/DDQN exhibits larger fluctuations. On the CS side, M/f-CS remains substantially lower and more volatile because its host decisions are random. Table~\ref{tab3} further shows that removing GAT preserves workflow success but increases energy consumption, while removing the learned CS policy reduces success to 92\% and increases energy to $2.19\times10^{7}$~J. These results indicate that the dependency-aware representation and learned host placement contribute in complementary ways: the former improves task-decision quality, and the latter coordinates locality and resource consolidation.



\subsection{Performance Comparison}

\subsubsection{Benchmark Settings}
We compare GA-HRL with five baselines representing the approaches of learning, heuristic, and optimization: DTODRL~\cite{10463608}, OHDS~\cite{9991105}, SMWDSA~\cite{9491116}, HACPPO~\cite{jayanetti2022deep}, and DS-CSP~\cite{hahnel2018extending}.
OHDS includes its own container-placement strategy, which is retained. For baselines that do not define host placement, we keep their original task policy and select the second-level placement randomly from feasible hosts. Within each run, all methods receive the same workflow instances, arrival process, and random seed.

\subsubsection{Effect of the deadline factor}
\begin{figure}[!t]
\centering
\includegraphics[width=3.4 in]{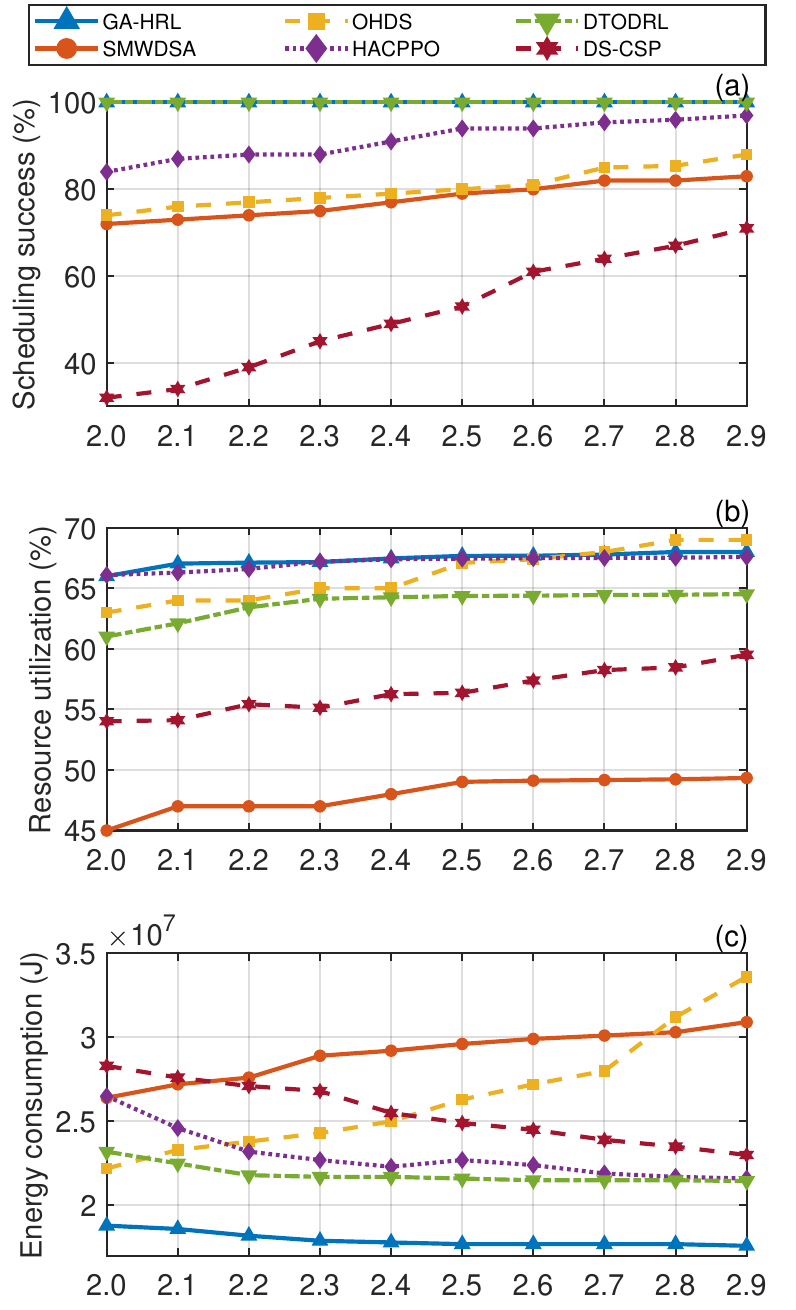}
\caption{Effect of the deadline factor $\alpha^{df}$ with $K=100$ workflows, $\lambda=0.5$, and $v=0.1$: (a) workflow scheduling success rate, (b) average container resource utilization, and (c) total system energy consumption.}
\label{fig_5}
\end{figure}

We first evaluate the effect of the deadline factor $\alpha^{df}$ with $K=100$ workflows, Poisson arrival rate $\lambda=0.5$, and execution-speed variation coefficient $v=0.1$. The results are shown in Fig.~\ref{fig_5}.

As shown in Fig.~\ref{fig_5}(a), relaxing the deadline generally enhances workflow scheduling success. GA-HRL and DTODRL maintain a 100\% success rate across the entire tested range. HACPPO also exhibits steady improvement with increasing $\alpha^{df}$ and consistently outperforms the heuristic and deterministic baselines, though it falls short of the two top-performing learning-based methods. In contrast, OHDS, SMWDSA, and especially DS-CSP yield lower success rates, as their policies are less responsive to the combined effects of dynamic arrivals, precedence constraints, and execution-speed variations.

The utilization and energy metrics in Fig.~\ref{fig_5}(b)--(c) offer a clearer distinction among the learning-based approaches. GA-HRL achieves the highest and most stable container utilization, ranging from 66\% to 68\%, at an energy cost of only $1.76\times10^7$-$1.88\times10^7$~J. DTODRL attains the same workflow success rate but utilizes only 61\%-64\% of container capacity and consumes $2.14\times10^7$-$2.32\times10^7$~J. HACPPO shows utilization close to that of GA-HRL yet incurs consistently higher energy consumption, indicating that similar container occupancy does not necessarily imply equivalent placement efficiency. Among the major baselines, SMWDSA remains the least resource-efficient, with utilization below 50\% and energy consumption around $3.09\times10^7$~J. Overall, these results suggest that GA-HRL can satisfy both loose and stringent deadlines without resorting to indiscriminate resource over-provisioning.

\subsubsection{The influence of execution-speed variation}
\begin{figure}[!t]
\centering
\includegraphics[width=3.4 in]{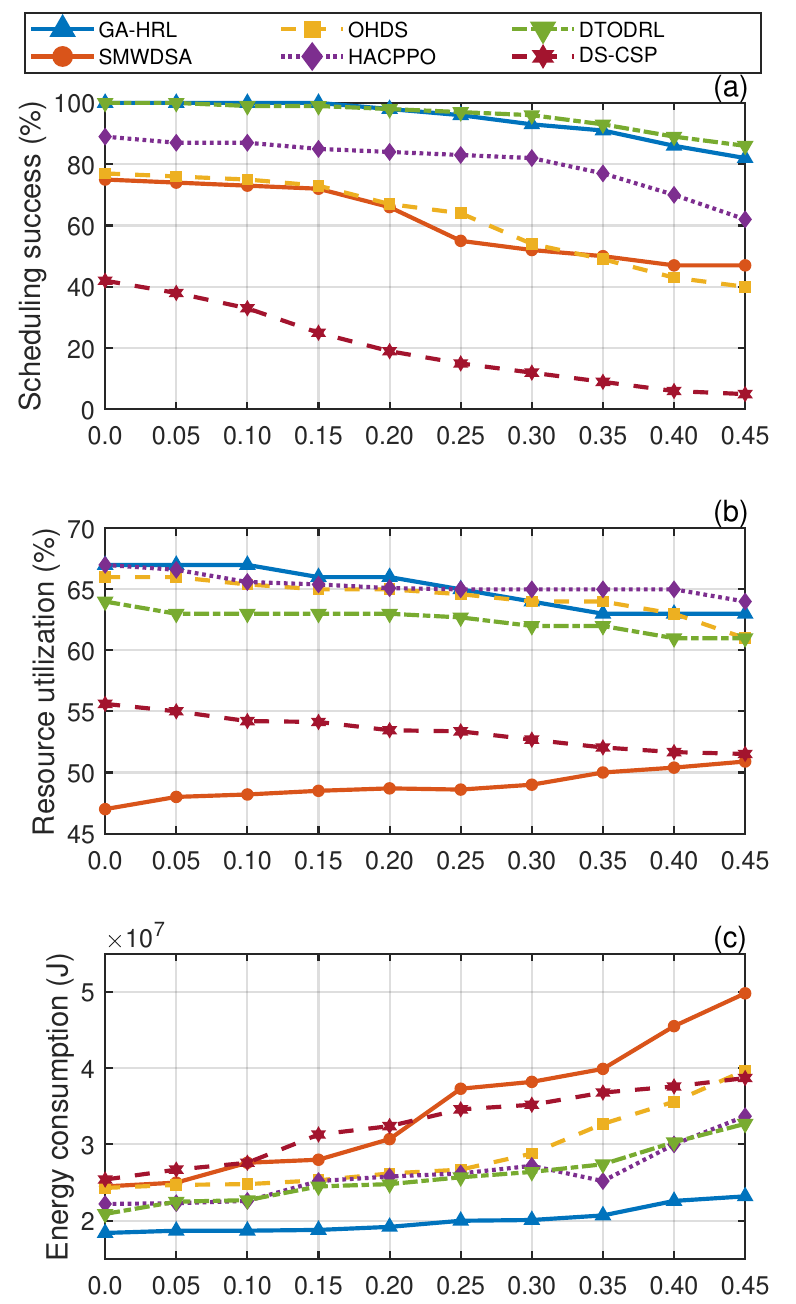}
\caption{Effect of the execution-speed variation coefficient $v$ with $K=100$ workflows, $\lambda=0.5$, and $\alpha^{df}=2.1$: (a) workflow scheduling success rate, (b) average container resource utilization, and (c) total system energy consumption.}
\label{fig_7}
\end{figure}

We next evaluate robustness to execution-speed variation by changing $v$ from 0 to 0.45 while fixing $K=100$, $\lambda=0.5$, and $\alpha^{df}=2.1$. The results are shown in Fig.~\ref{fig_7}.

Figure~\ref{fig_7}(a) shows that higher execution-speed variation lowers success for all methods, though at markedly different rates. At $v=0.45$, DTODRL leads with 86\%, followed by GA-HRL at 82\%. HACPPO exhibits intermediate robustness, degrading more slowly than heuristic baselines yet faster than the top two. In contrast, SMWDSA and DS-CSP drop to 47\% and 5\%, respectively, highlighting the fragility of reactive or deterministic policies under high variability.

The resource costs (Fig.~\ref{fig_7}(b)-(c)) reveal that GA-HRL maintains utilization at 63\%-67\% and energy at $2.32\times10^7$~J at $v=0.45$. DTODRL achieves 4\% higher success but consumes $3.27\times10^7$~J, while SMWDSA costs $4.98\times10^7$~J. HACPPO's utilization stays stable, yet its energy exceeds GA-HRL's with increasing volatility. Thus, at maximum variation, GA-HRL sacrifices a modest success margin (4\%) for roughly 29\% energy savings over DTODRL, which is consistent with its uncertainty-aware design that favors preserving deadline slack over aggressive scale-out.

\subsubsection{The influence of the number of workflows}
\begin{figure}[!ht]
\centering
\includegraphics[width=3.4 in]{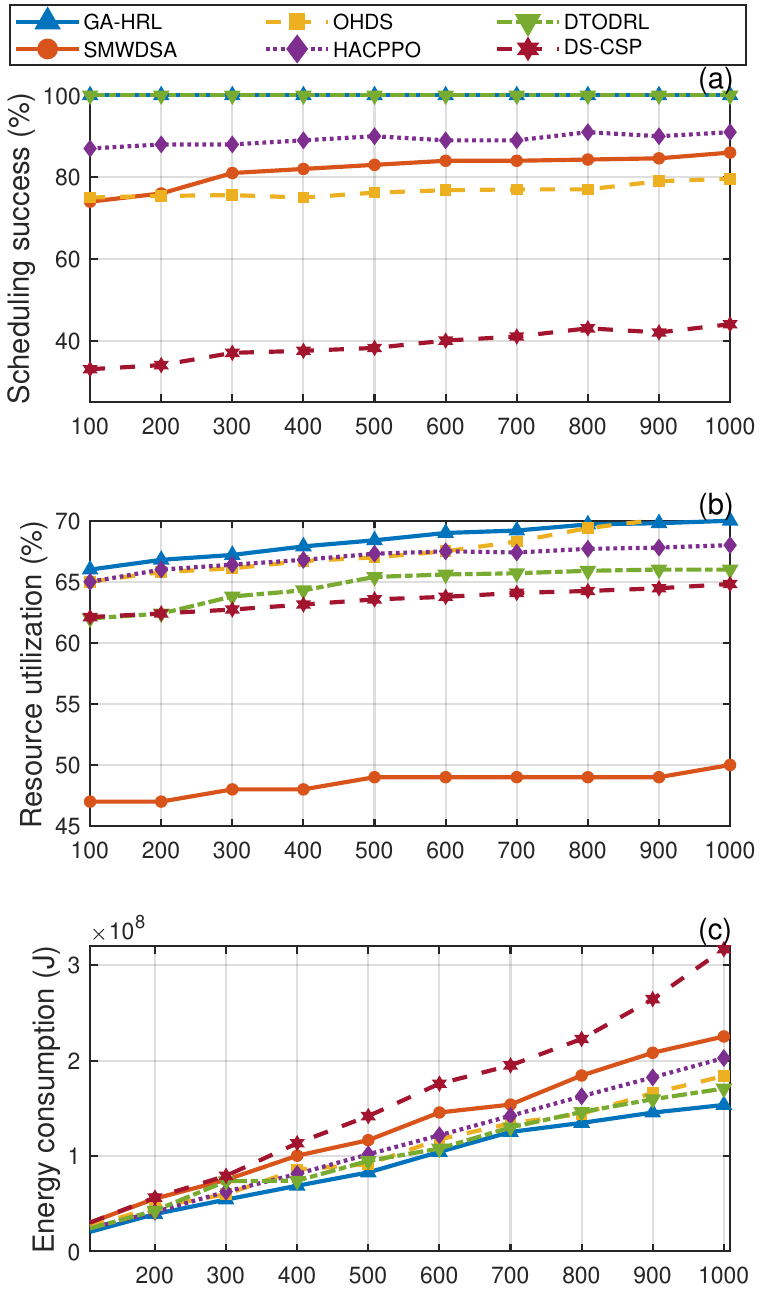}
\caption{Effect of the number of workflows $K$ with $\lambda=0.5$, $v=0.1$, and $\alpha^{df}=2.1$: (a) workflow scheduling success rate, (b) average container resource utilization, and (c) total system energy consumption.}
\label{fig_8}
\end{figure}

Finally, we evaluate scalability by varying the number of workflows $K$ from 100 to 1000 with $\lambda=0.5$, $v=0.1$, and $\alpha^{df}=2.1$. The results are shown in Fig.~\ref{fig_8}.

Fig.~\ref{fig_8}(a) shows that GA-HRL and DTODRL maintain workflow scheduling success close to 100\% as the workload increases. HACPPO remains comparatively stable in the high-success region but below GA-HRL and DTODRL, whereas OHDS and SMWDSA level off at lower success rates and DS-CSP degrades further. These results indicate that the learning-based schedulers are better able to absorb the denser arrival stream, but their resource efficiency differs.

As shown in Fig.~\ref{fig_8}(b), the average container utilization of GA-HRL rises gradually from about 66\% to 70\%, indicating that the scheduler increasingly reuses existing containers as more workflows overlap. HACPPO also maintains relatively high utilization, while DTODRL remains slightly lower than GA-HRL over most of the tested range. Fig.~\ref{fig_8}(c) shows that total energy consumption increases steadily with workflow volume for all methods. GA-HRL remains the most energy-efficient across the tested range and exhibits a more gradual increase than the competing schedulers. At $K=1000$, GA-HRL consumes $15.35\times10^{7}$~J, compared with $17.07\times10^{7}$~J for DTODRL and $22.53\times10^{7}$~J for SMWDSA; HACPPO also remains above GA-HRL in total energy. The results therefore support the conclusion that GA-HRL scales to denser workflow loads mainly through container reuse and coordinated placement rather than indiscriminate scale-out provisioning.

\section{Conclusion and Future Work}
\label{lab-sec-conclusion}



This paper presented GA-HRL, an event-driven hierarchical reinforcement learning scheduler for dynamic cloud workflows with stochastic execution speeds and placement-dependent communication. It represents workflows as DAGs, uses predicted sub-deadlines to capture task urgency, and applies a multi-head GAT to encode dependency information. A task-scheduling agent then assigns ready tasks to containers, and a container-scheduling agent places newly requested containers, with both agents trained alternately using separate PPO actor-critic networks. Trace-driven experiments on the 2018 Alibaba cluster trace showed that GA-HRL maintains competitive workflow success while generally achieving higher container utilization and lower energy consumption, and at the largest speed variation it trades a small success-rate gap relative to DTODRL for substantially lower energy. Ablation results further confirmed that the dependency-aware task representation and learned container placement improve scheduling efficiency in complementary ways. The current model represents runtime interference through execution-speed variation and does not explicitly consider resource failures or online estimation errors; future work will evaluate GA-HRL on a physical testbed and incorporate measured interference, failures, and online performance estimation.

\bibliographystyle{IEEEtran}
\bibliography{Mybib}

@ARTICLE{9491116,
  author={Ye, Lingjuan and Xia, Yuanqing and Yang, Liwen and Yan, Ce},
  journal={IEEE Transactions on Automation Science and Engineering}, 
  title={SHWS: Stochastic Hybrid Workflows Dynamic Scheduling in Cloud Container Services}, 
  year={2022},
  volume={19},
  number={3},
  pages={2620-2636},
  doi={10.1109/TASE.2021.3093341}}

@ARTICLE{9991105,
  author={Fan, Guisheng and Chen, Xingpeng and Li, Zengpeng and Yu, Huiqun and Zhang, Yingxue},
  journal={IEEE Transactions on Network and Service Management}, 
  title={An Energy-Efficient Dynamic Scheduling Method of Deadline-Constrained Workflows in a Cloud Environment}, 
  year={2023},
  volume={20},
  number={3},
  pages={3089-3103},
  doi={10.1109/TNSM.2022.3228402}}

@ARTICLE{8443134,
  author={Chen, Huangke and Zhu, Xiaomin and Liu, Guipeng and Pedrycz, Witold},
  journal={IEEE Transactions on Services Computing}, 
  title={Uncertainty-Aware Online Scheduling for Real-Time Workflows in Cloud Service Environment}, 
  year={2021},
  volume={14},
  number={4},
  pages={1167-1178},
  doi={10.1109/TSC.2018.2866421}}

@ARTICLE{10509784,
  author={Sun, Zaixing and Mei, Yi and Zhang, Fangfang and Huang, Hejiao and Gu, Chonglin and Zhang, Mengjie},
  journal={IEEE Transactions on Services Computing}, 
  title={Multi-Tree Genetic Programming Hyper-Heuristic for Dynamic Flexible Workflow Scheduling in Multi-Clouds}, 
  year={2024},
  volume={17},
  number={5},
  pages={2687-2703},
  doi={10.1109/TSC.2024.3394691}}

@ARTICLE{8000634,
  author={Wu, Quanwang and Ishikawa, Fuyuki and Zhu, Qingsheng and Xia, Yunni and Wen, Junhao},
  journal={IEEE Transactions on Parallel and Distributed Systems}, 
  title={Deadline-Constrained Cost Optimization Approaches for Workflow Scheduling in Clouds}, 
  year={2017},
  volume={28},
  number={12},
  pages={3401-3412},
  doi={10.1109/TPDS.2017.2735400}}

@article{jayanetti2022deep,
  title={Deep reinforcement learning for energy and time optimized scheduling of precedence-constrained tasks in edge--cloud computing environments},
  author={Jayanetti, Amanda and Halgamuge, Saman and Buyya, Rajkumar},
  journal={Future Generation Computer Systems},
  volume={137},
  pages={14--30},
  year={2022},
  publisher={Elsevier}
}

@ARTICLE{10497174,
  author={Liu, Zhang and Huang, Lianfen and Gao, Zhibin and Luo, Manman and Hosseinalipour, Seyyedali and Dai, Huaiyu},
  journal={IEEE Transactions on Network and Service Management}, 
  title={GA-DRL: Graph Neural Network-Augmented Deep Reinforcement Learning for DAG Task Scheduling Over Dynamic Vehicular Clouds}, 
  year={2024},
  volume={21},
  number={4},
  pages={4226-4242},
  doi={10.1109/TNSM.2024.3387707}}

@ARTICLE{10634877,
  author={Ding, Fan and Yuan, Yaqian and Lv, Lizhi and Zhang, Rui and Zhou, Wenbo},
  journal={IEEE Internet of Things Journal}, 
  title={Transformer-Enhanced DQN Approach for Energy and Cost-Efficient Large-Scale Dynamic Workflow Scheduling in Heterogeneous Environment}, 
  year={2024},
  volume={11},
  number={22},
  pages={37351-37367},
  doi={10.1109/JIOT.2024.3442997}}

@ARTICLE{10499978,
  author={Al Qassem, Lamees M. and Stouraitis, Thanos and Damiani, Ernesto and Elfadel, Ibrahim M.},
  journal={IEEE Transactions on Network and Service Management}, 
  title={Containerized Microservices: A Survey of Resource Management Frameworks}, 
  year={2024},
  volume={21},
  number={4},
  pages={3775-3796},
  doi={10.1109/TNSM.2024.3388633}}

@ARTICLE{9590522,
  author={Xie, Yanghao and Huang, Lin and Kong, Yuyang and Wang, Sheng and Xu, Shizhong and Wang, Xiong and Ren, Jing},
  journal={IEEE Transactions on Network and Service Management}, 
  title={Virtualized Network Function Forwarding Graph Placing in SDN and NFV-Enabled IoT Networks: A Graph Neural Network Assisted Deep Reinforcement Learning Method}, 
  year={2022},
  volume={19},
  number={1},
  pages={524-537},
  doi={10.1109/TNSM.2021.3123460}}

@ARTICLE{10061217,
  author={Yu, Xiaoming and Wu, Wenjun and Wang, Yangzhou},
  journal={IEEE Transactions on Services Computing}, 
  title={Integrating Cognition Cost With Reliability QoS for Dynamic Workflow Scheduling Using Reinforcement Learning}, 
  year={2023},
  volume={16},
  number={4},
  pages={2713-2726},
  doi={10.1109/TSC.2023.3253182}}

@ARTICLE{10463608,
  author={Cao, Zequn and Deng, Xiaoheng and Yue, Sheng and Jiang, Ping and Ren, Ju and Gui, Jinsong},
  journal={IEEE Internet of Things Journal}, 
  title={Dependent Task Offloading in Edge Computing Using GNN and Deep Reinforcement Learning}, 
  year={2024},
  volume={11},
  number={12},
  pages={21632-21646},
  doi={10.1109/JIOT.2024.3374969}}

@article{velickovic2017graph,
  title={Graph attention networks},
  author={Veli{\v{c}}kovi{\'c}, Petar and Cucurull, Guillem and Casanova, Arantxa and Romero, Adriana and Lio, Pietro and Bengio, Yoshua},
  journal={arXiv preprint arXiv:1710.10903},
  year={2017}
}

@misc{alibaba,
  year={Website. Alibaba Inc. (2018), [Online]. Alibaba Production Cluster Data v2018. Available: https://github.com/alibaba/clusterdata/tree/v2018}
}

@article{hahnel2018extending,
  title={Extending the cutting stock problem for consolidating services with stochastic workloads},
  author={H{\"a}hnel, Markus and Martinovic, John and Scheithauer, Guntram and Fischer, Andreas and Schill, Alexander and Dargie, Waltenegus},
  journal={IEEE Transactions on Parallel and Distributed Systems},
  volume={29},
  number={11},
  pages={2478--2488},
  year={2018},
  publisher={IEEE}
}

@inproceedings{das2020performance,
  title={Performance optimization for edge-cloud serverless platforms via dynamic task placement},
  author={Das, Anirban and Imai, Shigeru and Patterson, Stacy and Wittie, Mike P},
  booktitle={2020 20th IEEE/ACM International Symposium on Cluster, Cloud and Internet Computing (CCGRID)},
  pages={41--50},
  year={2020},
  organization={IEEE}
}

@article{mnih2015human,
  title={Human-level control through deep reinforcement learning},
  author={Mnih, Volodymyr and Kavukcuoglu, Koray and Silver, David and Rusu, Andrei A and Veness, Joel and Bellemare, Marc G and Graves, Alex and Riedmiller, Martin and Fidjeland, Andreas K and Ostrovski, Georg and others},
  journal={nature},
  volume={518},
  number={7540},
  pages={529--533},
  year={2015},
  publisher={Nature Publishing Group}
}

@inproceedings{jiao2017joint,
  title={Joint virtual network function selection and traffic steering in telecom networks},
  author={Jiao, Shundan and Zhang, Xiaoning and Yu, Shui and Song, Xue and Xu, Zhichao},
  booktitle={GLOBECOM 2017-2017 IEEE Global Communications Conference},
  pages={1--7},
  year={2017},
  organization={IEEE}
}

@article{deng2021dependent,
  title={Dependent function embedding for distributed serverless edge computing},
  author={Deng, Shuiguang and Zhao, Hailiang and Xiang, Zhengzhe and Zhang, Cheng and Jiang, Rong and Li, Ying and Yin, Jianwei and Dustdar, Schahram and Zomaya, Albert Y},
  journal={IEEE Transactions on Parallel and Distributed Systems},
  volume={33},
  number={10},
  pages={2346--2357},
  year={2021},
  publisher={IEEE}
}

@inproceedings{liu2019dependent,
  title={Dependent task placement and scheduling with function configuration in edge computing},
  author={Liu, Liuyan and Tan, Haisheng and Jiang, Shaofeng H-C and Han, Zhenhua and Li, Xiang-Yang and Huang, Hong},
  booktitle={Proceedings of the International Symposium on Quality of Service},
  pages={1--10},
  year={2019}
}

@article{cheng2015energy,
  title={An energy-saving task scheduling strategy based on vacation queuing theory in cloud computing},
  author={Cheng, Chunling and Li, Jun and Wang, Ying},
  journal={Tsinghua Science and Technology},
  volume={20},
  number={1},
  pages={28--39},
  year={2015},
  publisher={TUP}
}

@article{topcuoglu2002performance,
  title={Performance-effective and low-complexity task scheduling for heterogeneous computing},
  author={Topcuoglu, Haluk and Hariri, Salim and Wu, Min-You},
  journal={IEEE transactions on parallel and distributed systems},
  volume={13},
  number={3},
  pages={260--274},
  year={2002},
  publisher={IEEE}
}

@article{rodriguez2014deadline,
  title={Deadline based resource provisioningand scheduling algorithm for scientific workflows on clouds},
  author={Rodriguez, Maria Alejandra and Buyya, Rajkumar},
  journal={IEEE transactions on cloud computing},
  volume={2},
  number={2},
  pages={222--235},
  year={2014},
  publisher={IEEE}
}

@article{arabnejad2017scheduling,
  title={Scheduling deadline constrained scientific workflows on dynamically provisioned cloud resources},
  author={Arabnejad, Vahid and Bubendorfer, Kris and Ng, Bryan},
  journal={Future Generation Computer Systems},
  volume={75},
  pages={348--364},
  year={2017},
  publisher={Elsevier}
}

@article{yang2024scheduling,
  title={Scheduling workflow tasks with unknown task execution time by combining machine-learning and greedy-optimization},
  author={Yang, Yuanhao and Shen, Hong and Tian, Hui},
  journal={IEEE Transactions on Services Computing},
  volume={17},
  number={3},
  pages={1181--1195},
  year={2024},
  publisher={IEEE}
}

@article{jin2020review,
  title={A review of power consumption models of servers in data centers},
  author={Jin, Chaoqiang and Bai, Xuelian and Yang, Chao and Mao, Wangxin and Xu, Xin},
  journal={applied energy},
  volume={265},
  pages={114806},
  year={2020},
  publisher={Elsevier}
}

@article{kheldoun2017formal,
  title={Formal verification of complex business processes based on high-level Petri nets},
  author={Kheldoun, Ahmed and Barkaoui, Kamel and Ioualalen, Malika},
  journal={Information Sciences},
  volume={385},
  pages={39--54},
  year={2017},
  publisher={Elsevier}
}

@article{schulman2017ppo,
  author  = {Schulman, John and Wolski, Filip and Dhariwal, Prafulla and Radford, Alec and Klimov, Oleg},
  title   = {Proximal Policy Optimization Algorithms},
  journal = {arXiv preprint arXiv:1707.06347},
  year    = {2017}
}

\end{document}